\documentclass[journal]{IEEEtran}
\usepackage{graphicx}
\usepackage{enumerate}
\usepackage{amsmath,amsfonts,amssymb}
\usepackage{color}
\usepackage{multirow}
\usepackage[noend]{algpseudocode}
\usepackage{algorithmicx,algorithm}
\usepackage{setspace}
\usepackage{lineno,hyperref}
\usepackage{booktabs}
\usepackage[table]{xcolor}
\usepackage{xfrac}
\modulolinenumbers[5]
\usepackage{etoolbox}

\begin{document}

\title{SwiMDiff: Scene-wide Matching Contrastive Learning with Diffusion Constraint for Remote Sensing Image}

\author{
	Jiayuan Tian, Jie Lei,~\IEEEmembership{Member,~IEEE}, Jiaqing Zhang,
	Weiying Xie,~\IEEEmembership{Senior Member,~IEEE},
	Yunsong Li,~\IEEEmembership{Member,~IEEE}
\thanks{This work was supported in part by the National Natural Science Foundation of China under Grant 62071360, and in part by the Innovation Fund of Xidian University under Grant YJSJ2302. (\emph{Corresponding~authors: Jie Lei; Weiying Xie.})

Jiayuan Tian, Jie Lei, Jiaqing Zhang, Weiying Xie, Yunsong Li are with the State Key Laboratory of Integrated Services Networks, Xidian University, Xi’an 710071,
China (e-mail: jytian0414@stu.xidian.edu.cn; jielei@mail.xidian.edu.cn; jqzhang\underline{ }2@stu.xidian.edu.cn; wyxie@xidian.edu.cn; ysli@mail.xidian.edu.cn).

}
}

\markboth{Journal of \LaTeX\ Class Files,~Vol.~14, No.~8, November~2021}%
{Tian \MakeLowercase{\textit{et al.}}: SwiMDiff: Scene-wide Matching Contrastive Learning with Diffusion Constraint for Remote Sensing Image}

\maketitle
\begin{abstract}  

With recent advancements in aerospace technology, the volume of unlabeled remote sensing image (RSI) data has increased dramatically. Effectively leveraging this data through self-supervised learning (SSL) is vital in the field of remote sensing. However, current methodologies, particularly contrastive learning (CL), a leading SSL method, encounter specific challenges in this domain. Firstly, CL often mistakenly identifies geographically adjacent samples with similar semantic content as negative pairs, leading to confusion during model training. Secondly, as an instance-level discriminative task, it tends to neglect the essential fine-grained features and complex details inherent in unstructured RSIs.
To overcome these obstacles, we introduce SwiMDiff, a novel self-supervised pre-training framework designed for RSIs. SwiMDiff employs a scene-wide matching approach that effectively recalibrates labels to recognize data from the same scene as false negatives. This adjustment makes CL more applicable to the nuances of remote sensing. Additionally, SwiMDiff seamlessly integrates CL with a diffusion model. Through the implementation of pixel-level diffusion constraints, we enhance the encoder's ability to capture both the global semantic information and the fine-grained features of the images more comprehensively.
Our proposed framework significantly enriches the information available for downstream tasks in remote sensing. Demonstrating exceptional performance in change detection and land-cover classification tasks, SwiMDiff proves its substantial utility and value in the field of remote sensing.

\end{abstract}
\section{Introduction}
\IEEEPARstart{R}{EMOTE}
sensing image (RSI) analysis and interpretation hold paramount significance in the domain of computer vision, encompassing a range of distinct tasks such as land-cover classification \cite{gao3,gao4}, change detection \cite{gao1,shengao1,shengao2}, object detection \cite{gao2,ghost,superyolo}, etc. Such analysis facilitates the monitoring of natural phenomena and human activities on Earth's surface, encompassing domains like land-use surveillance \cite{sader1990remote}, disaster prevention \cite{schumann2018assisting}, precision agriculture \cite{mulla2013twenty}, and wildfire detection \cite{filipponi2019exploitation}. By capturing both natural occurrences and human-induced activities, RSI plays an indispensable role in applications spanning geographic information systems, agriculture, environmental science, and myriad other fields.

In the past decades, with the surge in aerospace technology, earth observation satellites generate terabytes of RSIs daily \cite{berg2022self}. Despite this abundance, two primary challenges persist: (1) High Specialist Manpower Requirement: The identification and labelling of RSIs necessarily demand professional researchers, resulting in high costs. (2) The Presence of Noisy Labels: The intrinsic complexity of RSIs makes generating flawless labels during large-scale data annotation challenging. This abundance of large but noisy labels is harmful for many tasks \cite{wang2022self}. Addressing these issues, the remote sensing community has shifted focus to automatic feature extraction and analysis from unlabeled RSIs \cite{peng1,peng2,peng3}. Self-supervised learning (SSL), exploiting the intrinsic structure of data, emerges as a key method to harness the potential of large-scale unlabeled RSIs. 


Early SSL methods largely relied on various pretext tasks \cite{cookbook}, such as jigsaw puzzles \cite{jigsaw}, patch localization \cite{patch}, and image inpainting \cite{inpainting}. These methods exhibit limited generalizability and are far surpassed by contrastive learning (CL). CL enhances representation by drawing similar instances closer and distancing dissimilar ones \cite{cl1, moco, mocov2, cl2, cl3byol,simclr,sim2}. It captures feature representations with high discriminability and strong generalizability, standing out among various SSL methods.

\begin{figure*}[htbp]
    \centering
   \includegraphics[width=0.88\textwidth]{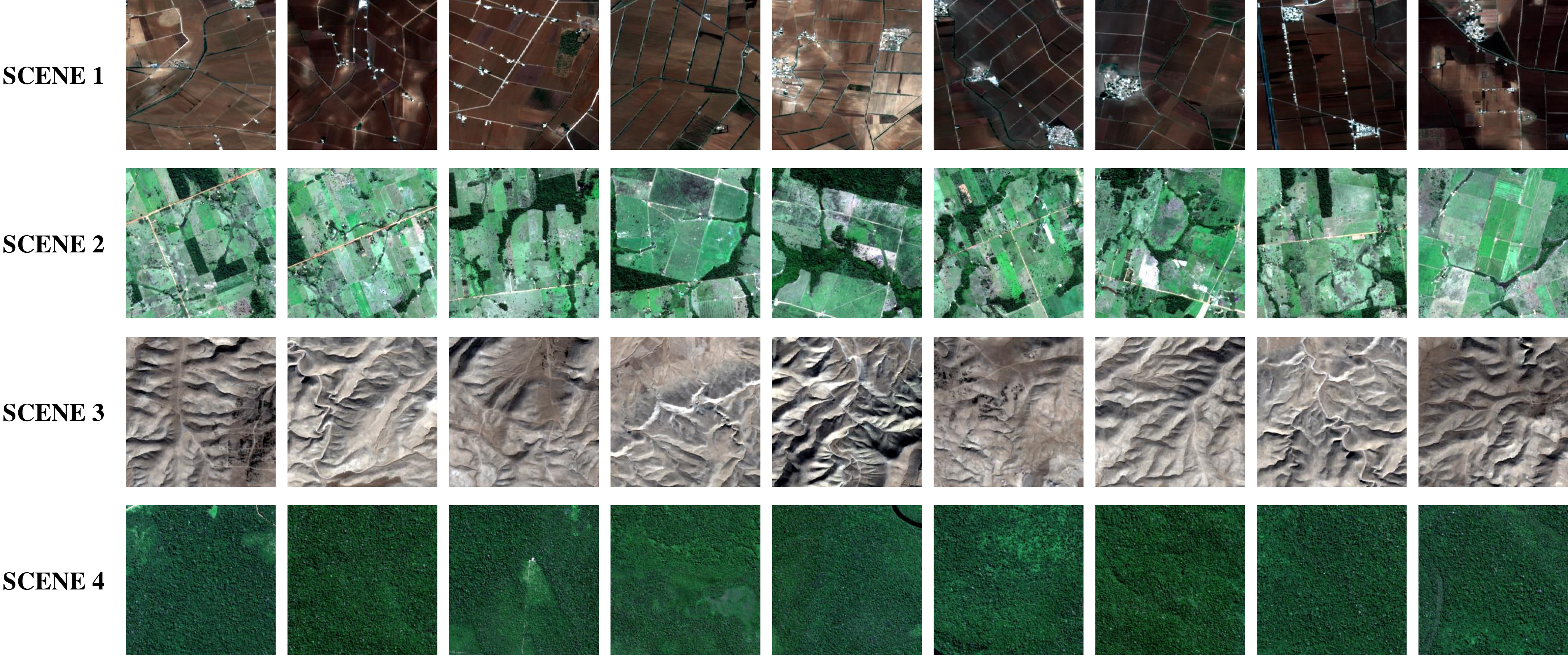}
    \caption{In remote sensing datasets, images are typically cropped from large scene images. Images cropped from the same scene exhibit certain similarities in terms of color, texture details, and overall layout.}
    \label{fig:1}
\end{figure*}

However, efficiently applying CL in remote sensing is hindered by two main obstacles. First, as delineated by the fundamental laws of geography \cite{Tobler}, data samples with close geographical proximity should inherently exhibit a degree of similarity. As depicted in Fig. \ref{fig:1}, images from the same scene demonstrate significant semantic and perceptual similarities. However, the current CL paradigm tends to classify geographically and semantically similar samples as negatives, overlooking their potential mutual connections and resulting in sample confusion \cite{false}.
Second, RSIs often lack clear foreground-background distinctions, with key information randomly distributed throughout the entire image. However, CL, as a global discriminant task, excels in extracting global discriminative information and inherently struggles to capture details. This inherent challenge makes it difficult for CL to capture the fine-grained details essential for remote sensing tasks.
To overcome these limitations, we propose SwiMDiff, a novel self-supervised pre-training framework tailored for RSIs. SwiMDiff enhances CL by incorporating a scene-wide matching strategy. This strategy recalibrates data based on the associated scenes to avoid mislabeling samples as negatives. It incorporates consideration for intra-class similarity, enhancing the model's ability to extract global discriminative and semantic information.
Furthermore, by integrating a diffusion model, SwiMDiff strengthens the CL encoder, creating a unified training framework that synergizes both techniques. Empowered by pixel-level diffusion constraints, SwiMDiff places a greater emphasis on local and detailed information. SwiMDiff not only captures global semantic information based on CL but also pays more attention to fine-grained features, enabling it to capture and provide richer information for downstream tasks in the remote sensing domain.

Our evaluation of SwiMDiff involves change detection and land-cover classification tasks on datasets like OSCD \cite{oscd} and LEVIR-CD \cite{levir} for change detection, and BigEarthNet \cite{bigearthnet} and EuroSAT \cite{EuroSAT} for land-cover classification. The results demonstrate SwiMDiff's superior performance, indicating its potential promising applicability in the realm of remote sensing.

In summary, this paper makes the following contributions:
\begin{itemize}
    \item 
 The development of SwiMDiff, an innovative self-supervised pre-training framework for RSI analysis. This framework is pioneering in its integration of the diffusion model with CL training, establishing a new benchmark in effective pre-training methodologies for RSI.
    \item 
The implementation of a scene-wide matching strategy within the SwiMDiff framework. This approach capitalizes on the semantic similarities of images from identical geographical scenes. It significantly enhances SwiMDiff's ability to accurately interpret and analyze RSIs.
    \item 
Demonstrated excellence in practical applications. SwiMDiff has been rigorously tested on change detection tasks using the OSCD and LEVIR-CD datasets and on land-cover classification tasks with the BigEarthNet and EuroSAT datasets. In these applications, SwiMDiff has achieved state-of-the-art results, showcasing its exceptional capacity to handle a variety of complex remote sensing tasks.
\end{itemize}

\section{Related Work}
\label{sec:Related Work}
\subsection{Diffusion Model}
Diffusion models, a subset of probabilistic denoising networks, operate by introducing noise into data, then learning to reverse this process to generate refined samples \cite{diffzongshu}. A notable subclass of these models, the Denoising Diffusion Probabilistic Models (DDPMs) \cite{diff}, encompass forward and reverse mechanisms. In the forward process, random noise is incrementally introduced to an image, functioning like a Markov chain where each step depends only on its predecessor. Over time, this added noise transforms the original image into a purely noisy state. Conversely, the reverse process uses the noise patterns from the forward process to reconstruct the original image. This is achieved by feeding the noisy image into a neural network, which is trained to predict the noise, constrained by the $\ell _{2}$-norm between the predicted and actual noise.

Diffusion models have rapidly advanced in recent years, showing exceptional promise in fields like computer vision \cite{diffgene1,diffgene2,ramesh2022hierarchical}. Their generative capabilities have been leveraged for robust feature extraction and representation. For instance, Xiang \emph{et al.} utilized the networks within diffusion models for self-supervised feature extraction \cite{d}. In remote sensing, Bandara \emph{et al.} applied DDPM for SSL on RSIs, leading to applications in change detection \cite{ddpm}. Another application in remote sensing is the BSDM approach, which employs diffusion models' denoising and representational strengths for background suppression in hyperspectral images, facilitating hyperspectral anomaly detection tasks \cite{mabo}. Hence, we consider the diffusion model for self-supervised representation learning. However, relying solely on it demands substantial data, a high-parameter network, and extensive iterations \cite{d}, struggling to capture global semantic information. Therefore, we propose leveraging the diffusion model to assist CL, enhancing the encoder for more accurate advanced semantics and richer fine-grained details.

\subsection{False Negative Samples in Contrastive Learning}
Since Hadsell \emph{et al.} introduced the CL paradigm \cite{hadsell}, it has continuously been optimized and improved, centering around the core concept of positive and negative samples. CMC leverages varying views of the same object to enhance mutual information \cite{cmc}. MoCo \cite{mocov2} and SimCLR \cite{sim2} distinguish augmented samples within momentum-updated queues and memory banks, respectively, for pretext task of instance discrimination, significantly enhancing the efficacy and practicality of CL. Conventionally, different augmentations of the same image are considered positive pairs, while different images are treated as negatives. However, this straightforward classification can lead to the creation of false negative samples (FNSs), where similar-category samples are erroneously marked as negatives. This mislabeling causes sample confounding, impairing model accuracy \cite{false}.

Some studies eliminate negative samples. For example, BYOL \cite{cl3byol} and SimSiam \cite{simsiam} solely utilize the similarity between positive pairs for predictive tasks, and Barlow Twins aims to approximate the cross-correlation matrix generated from positive pairs to the identity matrix \cite{cl4barlow}. However, their effectiveness has not seen substantial improvement, failing to fundamentally resolve the issue of FNS. Other studies focus on redefining sample relationships, effectively addressing the FNS problem. ASCL, another innovative method, adapts the traditional instance discrimination task into a multi-instance format by dynamically relabeling FNSs, thereby improving the overall efficacy of CL \cite{ascl}. FALSE method tackles FNSs in RSIs by adjusting their influence in the loss function, based on coarse and precise assessments \cite{false}. However, traversing the entire database for FNSs is resource-intensive and lacks close correlation with positive samples, resulting in suboptimal feature extraction. In this paper, our proposed SwiMDiff precisely locates and recalibrates FNSs within the same scene subset, significantly enhancing precision and efficiency in acquiring FNSs during CL.
\subsection{Self-supervised Learning in Remote Sensing}
The advancement and proliferation of airborne and satellite optical sensors have made remote sensing imagery more accessible than ever. This ease of access is further supported by large-scale remote sensing datasets, such as SEN12ms \cite{sen12ms} and fMoW \cite{fmow}, and platforms like Google Earth Engine \cite{google}, which facilitate the creation of customized datasets. These developments are significantly advancing SSL in the remote sensing field.

Remote sensing data's distinct characteristics have led to the development of specialized SSL methods. For instance, Ayush \emph{et al.} have utilized the spatiotemporal nature of RSIs to introduce a geographic location classification task in CL, effectively merging self-supervised and supervised learning techniques \cite{geossl}. SauMoCo has been developed to enhance semantic diversity by exploiting the semantic similarities between adjacent tiles in RSIs \cite{saumoco}. SeCo takes advantage of the seasonal consistency in RSIs to maximize the use of satellite imagery \cite{seco}. Additionally, STICL employs optimal transport techniques to handle RSIs across different spatial-temporal scenes, aiming to learn spatial-temporal invariant representations \cite{sticl}.



\begin{figure*}[tbp]
    \centering
   \includegraphics[width=0.9\textwidth]{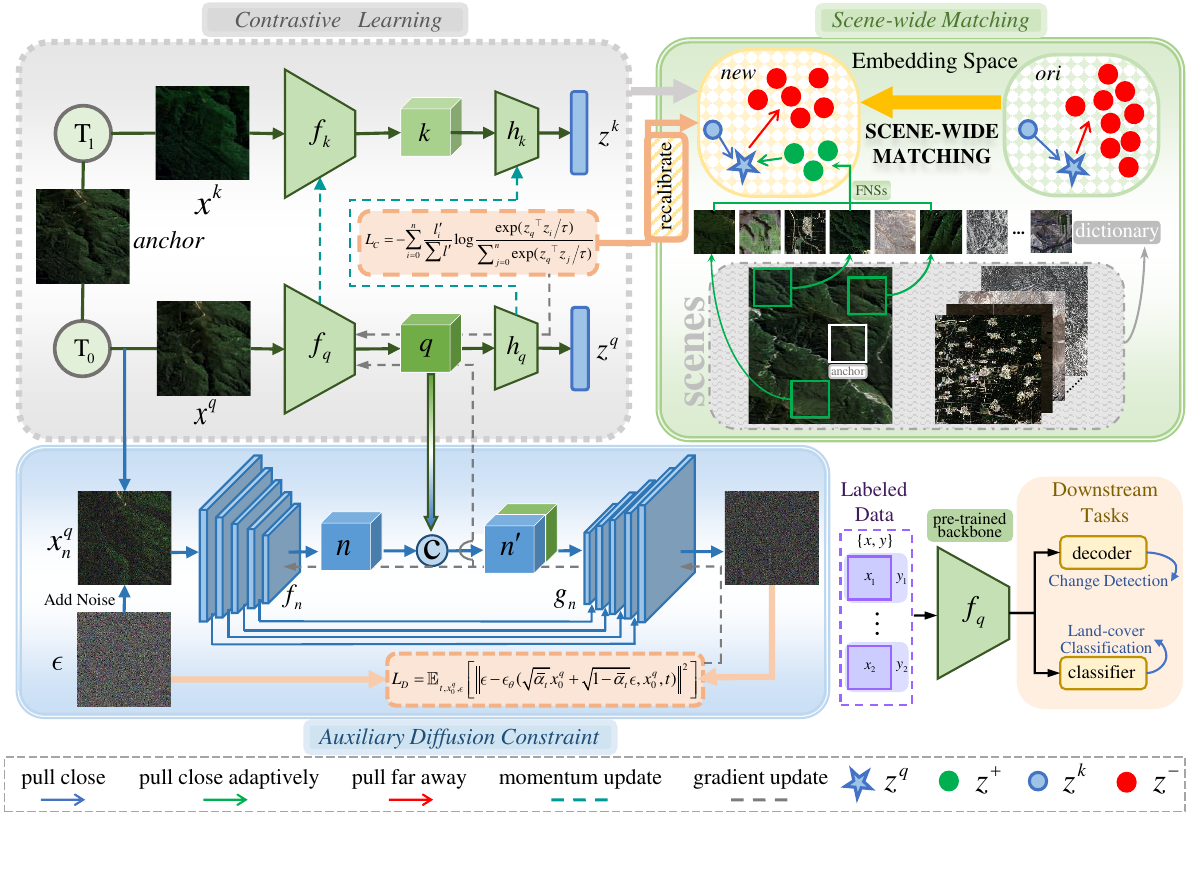}
    \caption{Diagram of the SwiMDiff. The network architecture is bifurcated into two components: 1) A dual-branch structure for CL. 2) A diffusion model network comprising an encoder and a decoder.}
    \label{fig:2}
\end{figure*}
\section{Method}
\label{sec:metod}
We propose SwiMDiff, a novel self-supervised pre-training framework for RSIs, combining the strengths of CL and generative learning. SwiMDiff aims to refine an encoder for enhanced feature extraction in various downstream tasks. As depicted in Fig. \ref{fig:2}, SwiMDiff comprises two main components: a dual-branch CL network (Section \ref{Section 3.1}) and a diffusion model network (Section \ref{Section 3.2}).
\subsection{Scene-wide Matching Contrastive Learning}
\label{Section 3.1}
To enhance representation extraction for RSIs, we introduce a scene-wide matching approach within the CL framework. This method builds upon the foundation of CL and incorporates the concept of scene-wide matching. The network architecture is similar to the original CL network, consisting of dual-branch momentum-updated encoders and projection heads. Our novel contribution lies in the embedding space, specifically recalibrating the negative sample set in the dictionary for more accurate loss computation.


\subsubsection{Contrastive Learning}
An anchor image is subjected to different augmentations $T_{0}$ and $T_{1}$, such as random cropping, color jittering, and flipping \cite{moco}, generating query view $x^{q}$ and key view $x^{k}$ for the CL network. The two views are processed by the network encoder $f_{q}$ and the momentum encoder $f_{k}$, yielding features $q= f_{q} (x^{q} )$ and $k= f_{k} (x^{k} )$. Then the projection heads $h_{q}$ and $h_{k}$ transform high-dimensional features into the representations $z_{q}=  h_{q} \left ( q \right )$ and $z_{k}=  h_{k} \left ( k \right )$ in a spherical space. The original loss seeks to minimize the distance between $z_{q}$ and $z_{k}$ while concurrently maximizing the separation from the dictionary queue. And we unify all samples as $\left \{ z_{k}, z_{1},\dots,z_{n} \right \} \triangleq \left \{ z_{0}, z_{1},\dots,z_{n} \right \}$:
\begin{equation}
\begin{split}
L_{ori}&= - \log_{}{}\frac{\exp \left ( {z_{q} }^\top z_{k}  \right /\tau ) }{\exp\left ( {z_{q} }^\top z_{k} /\tau\right ) + \sum_{i=1}^{n} \exp \left ( {z_{q} }^\top z_{i} /\tau\right )  }\\
&=-\sum_{i=0}^{n}l_{i}\log_{}{} \frac{\exp \left ( {z_{q} }^\top z_{i}  \right /\tau ) }{ \sum_{j=0}^{n} \exp \left ( {z_{q} }^\top z_{j} /\tau\right )  } , 
\end{split}
\end{equation}
where
$$l_{i} =
\begin{cases}
1,& i=0\\
0,& otherwise.
\end{cases}$$

Here ${z_{q} }^\top$ represents the transpose of $z_{q}$, and $\left \{ l_{0}, l_{1},\dots,l_{n} \right \}$ refers to the one-hot pseudo label showing that only $x^{k} $ belong to the same class as $x^{k}$.
\subsubsection{Scene-wide Matching}
As illustrated in Fig. \ref{fig:2}, images under a broad scene exhibit varying degrees of semantic similarity and intra-class relationships. We identify images belonging to the same scene with the anchor as FNSs. Initially, we select FNSs from the dictionary bank and constitute a false negative set $Z_{FNS}=\left \{ z_{1}^{+},  z_{2}^{+},\dots ,z_{m}^{+}\right \}  $. Conversely, the remaining samples are still maintained as negatives $z^{-}$. Given the intricate and multifaceted nature of the surface environment, each sample from $Z_{FNS}$ encompasses pertinent but not identical geographical information. Thus, we utilize an adaptive soft mechanism \cite{ascl} to re-formulate FNSs.

Based on the cosine similarity $\left \{ d_{1}, d_{2},\dots,d_{m} \right \}$ between FNSs and $z_{0}$, the relative distribution $\left \{ b_{1}, b_{2},\dots,b_{m} \right \}$ between $z_{0}$ and other representations in $Z_{FNS}$ can be considered as soft labels as:
\begin{equation}
    b_{i} =\frac{\exp (d_{i}/\tau ’ )}{\sum_{j=1}^{m}\exp (d_{i}/\tau ’ ) },i=1,\dots ,m.
\end{equation}

Subsequently, the adaptive soft labels are derived as $\left \{ s_{1}, s_{2},\dots,s_{m} \right \}$ and limited less than one for keeping the positive itself $z_{0}$ still the most confident sample:
\begin{equation}
    s_{i} =min(1,b_{i} \ast \left [   1-\frac{H\left ( \mathbf{b}  \right ) }{\log\left ( n \right )  } \right ]),i=1,\dots ,m,
\end{equation}
where $H\left ( \mathbf{b}  \right )$ represents the Shannon entropy. The scene-wide matching labels are relabeled as:
$$l'_{i} =
\begin{cases}
1,& i=0\\
s_{i},& i\neq 0 \cap z_{i}\in Z_{FNS}  \\
0,& otherwise.
\end{cases}$$

Here $\left \{ l'_{0}, l'_{1},\dots,l'_{n} \right \}$ indicates $z_{0}$ as the most confident positive. Other FNSs from the same scene are adaptively assigned their respective label values based on varying similarities. Conversely, other samples are still considered as unrelated negatives. Ultimately, we employ the normalized $l'$ to obtain the scene-wide matching contrastive loss:
\begin{equation}
    L_{C}= -\sum_{i=0}^{n}\frac{l'_{i}}{\sum l'} \log_{}{} \frac{\exp \left ( {z_{q} }^\top z_{i}  \right /\tau ) }{ \sum_{j=0}^{n} \exp \left ( {z_{q} }^\top z_{j} /\tau\right )  }  .
\end{equation}

Through the contrastive loss $L_{C}$, the network encoder $f_{q}$ is trained to distill global discriminative and deep semantic information.

\begin{figure}[tbp]
    \centering
    \includegraphics[width=0.45\textwidth]{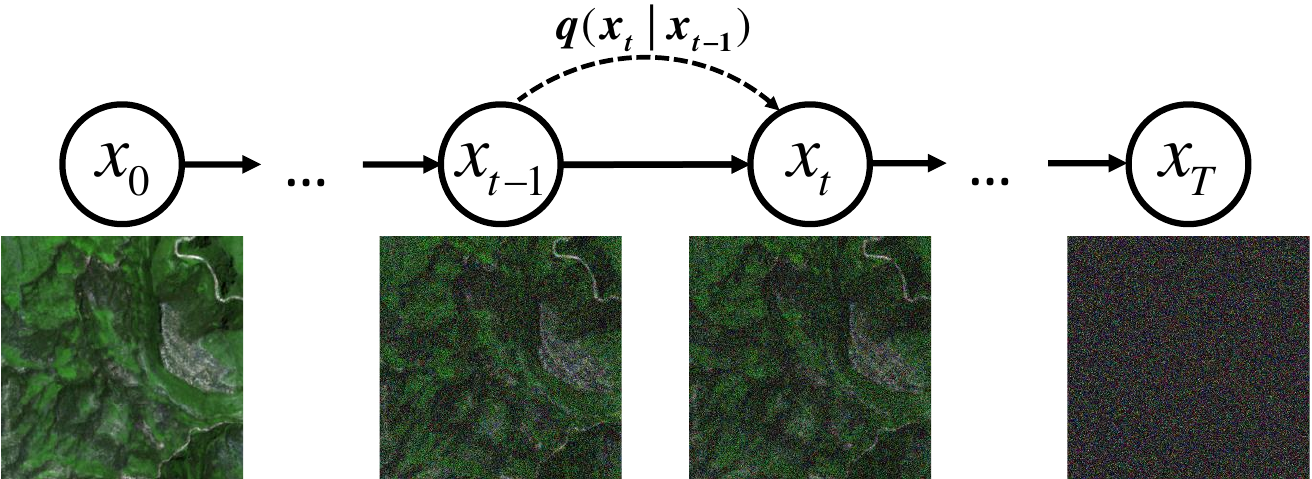}
    \caption{The forward diffusion process of the diffusion model. It's similar to a Markov chain, where noise is added based on the previous state.}
    \label{fig:3}
\end{figure}

\subsection{Auxiliary Diffusion Constraint} 
\label{Section 3.2}
To extract finer details from RSIs for downstream tasks, we integrate a diffusion model as an auxiliary task, adding a diffusion constraint to $f_{q}$ for iterative optimization. We adopt the diffusion model’s training mechanism, predicting noise introduced during the forward process with both encoder $f_{n}$ and decoder $g_{n}$.


\subsubsection{Gaussian Forward Diffusion}
\label{sssec:subsubhead}
We first add random Gaussian noise $\epsilon \sim \mathcal{N}   \left ( 0,  \boldsymbol{I}  \right ) $ to the clean image $x^{q}$ (i.e. $x_{0}^{q}$) $t$ times cumulatively. As depicted in Fig. \ref{fig:3}, the noising operation at step $t$ solely related to step $t-1$ is defined as follows:
\begin{equation}
    q\left ( x_{t }^{q}  |  x_{t-1 }^{q} \right ) =\mathcal{N } \left ( x_{t}^{q}; \sqrt{1-\beta _{t} }x_{t-1}^{q},\beta _{t} \boldsymbol{I}  \right ),
\end{equation}
where $\left ( x_{0}^{q}, x_{1}^{q},\dots ,x_{T}^{q} \right ) $ represents a Markov chain of noisy images and $\left ( \beta_{1}, \beta_{2},\dots ,\beta_{t} \right ) $ denotes the coefficients linearly interpolated between 0.0001 and 0.02, governing the noise variance at each step. By accumulating noise over $t$ iterations and reparameterizing the intermediate process, we then obtain the distribution of noising image as:
\begin{equation}
\begin{split}
    q\left ( x_{t}^{q} |x_{0}^{q}  \right )&=  \prod_{i=1}^{t}  q\left ( x_{i}^{q} |x_{i-1}^{q}  \right )\\
& =\mathcal{N}\left ( x_{t}^{q};\sqrt{\bar{\alpha }_{t}  } x_{0}^{q} ,\left ( 1- \bar{\alpha }_{t}\right )  \boldsymbol{I} \right )  ,
\end{split}
\end{equation}
where $\alpha _{t} =1-\beta _{t} $ and $\bar{\alpha } _{t} =\prod_{i=1}^{t} \alpha _{i} $. With $t$ growing, $x_{t}$ gets closer to pure Gaussian distribution. Based on Eq.(6), we derive $x_{n}^{q} $ by adding random $n$ step noise to $x_{0}^{q} $. 
\subsubsection{Training Mechanism}
\label{sssec:subsubhead}

The obtained noised image $x_{n}^{q} $ is used as input for the diffusion model's U-Net network. Through the encoder $f_{n} $, it is mapped to high-dimensional features $n=f_{n} \left ( x_{n}^{q}  \right)$.
Notably, the two networks are integrated by concatenating feature $q$ emerged from the encoder $f_{q}$ of CL and feature $n$ into ${n}'$, which is subsequently fed into the decoder $g_{n}$. The dimensions of the two feature maps are consistent, achieving integration at the feature level. 

The feature $q$ retains the global discriminative information and high-level semantic content of the clean image. It serves as guiding information for the diffusion model, facilitating noise prediction.
Guided by $q$, both $f_{n}$ and $g_{n}$ strive to predict the noise added to $x_{0}^{q} $. We condition the step estimation function $\epsilon$ with a noise-free image prior, denoted as:
\begin{equation}
    \begin{split}
        \epsilon _{\theta }\left ( x_{n}^{q} ,x_{0}^{q},t  \right )  &=g_{n}\left ( \left ( f_{n}\left ( x_{n}^{q},t \right ) \oplus f_{q}\left ( x_{0}^{q} \right ) \right ) ,t \right ) \\
    &= g_{n}\left ( \left ( n\oplus q \right ) ,t \right ) \\
    &=g_{n}\left ( n^{\prime}  ,t \right ),
    \end{split}
\end{equation}
where $\oplus$ denotes the concatenation operation of two feature matrices. Furthermore, $f_{q}$, $f_{n}$ and $g_{n}$ are trained to minimize the pixel-level distance between the original noise $ \epsilon_{\theta}$ and the predicted noise:
\begin{equation}
    L_{D}=\mathbb{E}_{t,x_{0}^{q},\epsilon  }  \left [ \left \| \epsilon -\epsilon _{\theta }\left ( \sqrt{ \bar{\alpha }_{t} } x_{0}^{q} + \sqrt{ 1-\bar{\alpha }_{t} }\epsilon,x_{0}^{q},t \right )    \right \|^{2}   \right ] .
\end{equation}

\begin{figure}[tbp]
    \centering
    \includegraphics[width=0.46\textwidth]{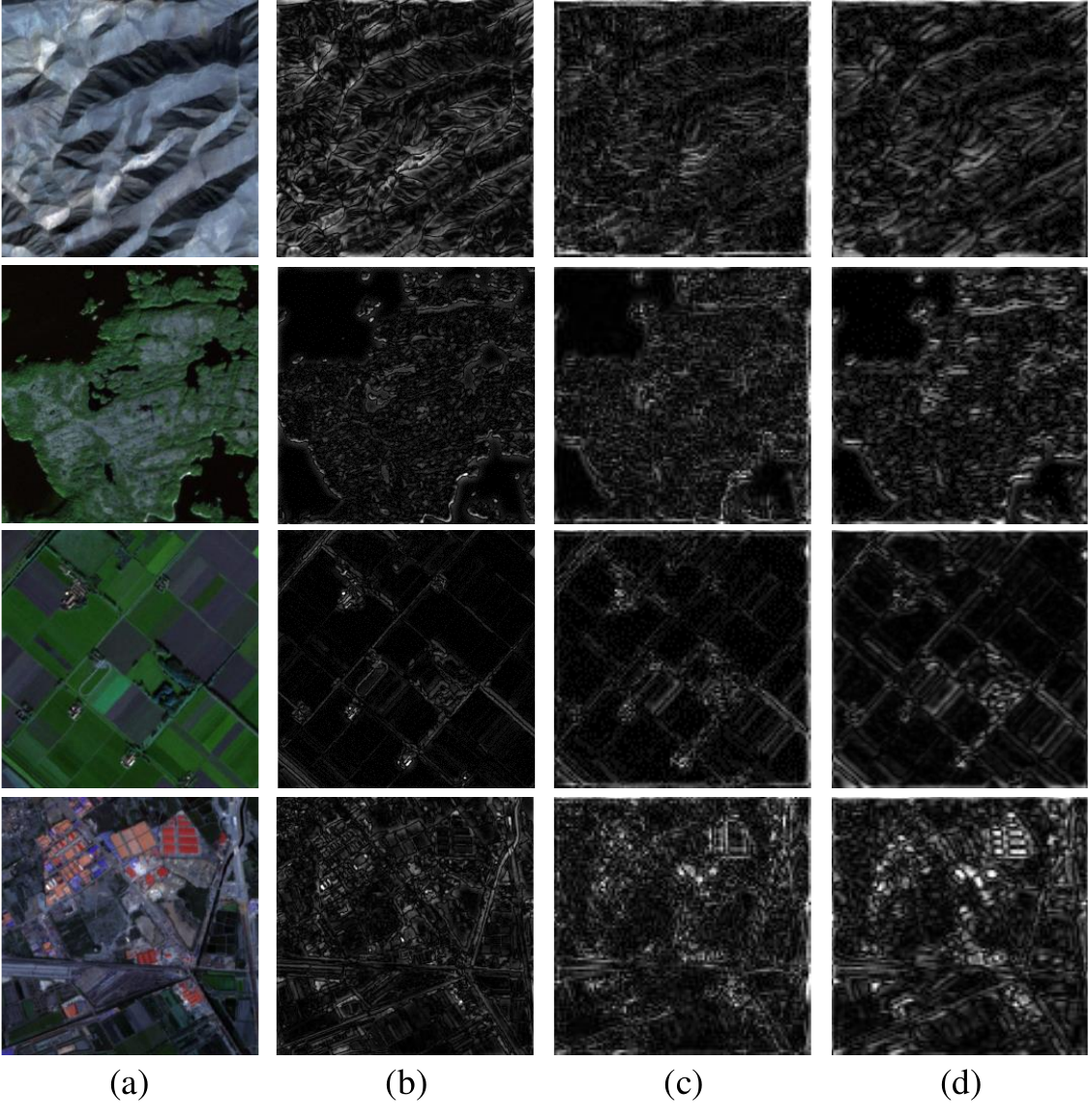}
    \caption{High-frequency components of the images and its shallow features. (a): Input Images. (b): High-frequency components extracted from input images. (c): High-frequency details of shallow features extracted from encoder pre-trained by CL. (d): High-frequency details of shallow features extracted from encoder pre-trained by model integrating with the diffusion model.}
    \label{high}
\end{figure}

Currently, by integrating CL and the diffusion model, it essentially establishes a multi-task learning framework. The addition of diffusion model-based pixel-level constraints renders the participating encoder $f_{q}$ to receive richer and more detailed supervisory signals. This results in the preservation and emphasis of shallow features, such as edges and textures, as shown in Fig. \ref{high}. We represent the fine-grained information contained in features by extracting high-frequency components. Compared to the original image, the features after integrating the diffusion model exhibit clearer boundaries and richer textural details. Fig. \ref{high} intuitively demonstrates that the diffusion model, serving as an auxiliary branch, can significantly reduce information loss in CL, providing a more effective and comprehensive feature representation for downstream tasks.

\subsection{Joint Training}
\label{Sec:Focus Removal}
The overall objective of SwiMDiff is formulated as:
\begin{equation}
    L=\lambda _{C}\ast L_{C}+  \lambda _{D}\ast L_{D}.
\end{equation}
Here, $\lambda _{C}$ and $\lambda _{D}$ are weight factors balancing the impacts of the two losses. SwiMDiff not only enhances discriminability among samples but also preserves detailed features within images. The scene-wide matching CL incorporates similarity information between adjacent tiles, improving high-level discriminative feature extraction. The diffusion constraint enables the encoder to grasp the data's inherent structural and distributional nuances, focusing on pixel-level details.

\section{Experimental Results}
\label{sec:Experiment}
We assess the representations produced by our method across two downstream tasks: change detection task and land-cover classification task.
\subsection{Self-supervised Pre-training}
\subsubsection{Pre-training Dataset} 
We utilize a subset of the Sen12MS \cite{sen12ms} dataset for pre-training, considering time constraints. The Sen12MS dataset comprises 180,662 triplets of Sentinel-1 SAR data, Sentinel-2 \cite{s2} multispectral images, and MODIS maps. The Sentinel-1 SAR data contains 2 spectral bands, while the Sentinel-2 multispectral image is comprised of 10 spectral bands. For our pre-training, we randomly select 10,000 atmospherically corrected Sentinel-2 RGB images (256x256) from each season, with 85\% allocated for training and 15\% allocated for validation.


\subsubsection{Pre-training Implementation Details}  
Our framework integrates the scene-wide matching method with \textbf{Momentum Contrast }(MoCo-v2) \cite{mocov2} baseline and employs the \textbf{DDPM} \cite{diff} framework for the diffusion model. We use ResNet-18 \cite{resnet} as our encoders and a 2-layer MLP as the projection head. The model is pre-trained for 1000 epochs with a batch size of 256 distributed across 4 Nvidia A100 GPUs. For the CL network, we employ an SGD optimizer, setting the initial learning rate at 0.03, the momentum at 0.9, and the weight decay at 1e-4. Meanwhile, the diffusion model network is trained using an Adam optimizer \cite{adam} with a learning rate of 1e-3. The joint training involves weight factors $\lambda_C$ of 1 and $\lambda_D$ of 10. Additionally, we set the temperature scaling parameters $\tau $ and $\tau '$ at 0.1 and 0.05, respectively, in the contrastive loss component.



\subsubsection{Comparative Methods}
We compare our proposed SwiMDiff with several methods, consisting of random initialization and some self-supervised pre-training. The self-supervised pre-training methods include MoCo-v2 \cite{mocov2} (baseline), Barlow Twins \cite{cl4barlow} (eliminating negative samples and reducing redundancy), DiRA \cite{dira} (uniting discriminative, restorative, and adversarial learning in a unified manner), and tri-SimCLR \cite{tri} (introducing a 3-factor contrastive loss).

Additionally, our evaluation extends to ablation studies focusing on specific components of our framework. This includes an examination of the impact of scene-wide matching when added to the baseline MoCo-v2 (termed MoCo-v2+SwiM) and an analysis of the joint pre-training approach that combines CL with the diffusion model (denoted as MoCo-v2+Diff). Each of these methods, along with our proposed SwiMDiff, is trained on the same 10,000-sample subset of the Sen12MS dataset.



\begin{figure}[tbp]
    \centering
    \includegraphics[width=0.48\textwidth]{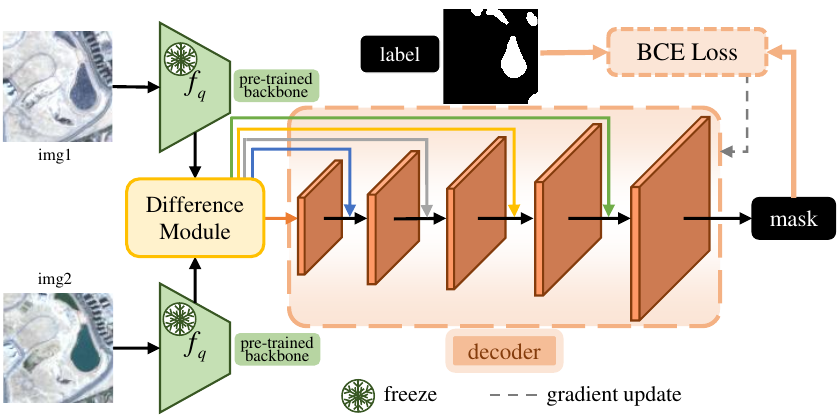}
    \caption{The network architecture for change detection task. The images taken at different times are first processed by a pre-trained and frozen encoder $f_{q}$. It extracts two sets of features from these images. These feature sets are then passed through a difference module. The resulting difference is then input into the decoder for further processing.}
    \label{change_kuangtu}
\end{figure}
\subsubsection{Metrics in Downstream Tasks}
We opt for change detection and land-cover classification as our downstream tasks, which are both significant and widely examined in remote sensing \cite{seco,shengao1,shengao2}. These two tasks provide different and valuable perspectives for evaluating the pre-trained model roundly, one emphasizing shallow-level details and the other focusing on deep-level semantics.

We employ various mathematical metrics to assess performance in change detection and land-cover classification. The outcomes of these downstream tasks reflect the performance of models obtained through different self-supervised methods.
\paragraph{Metrics in Change Detection}
In change detection, we utilize the F1 score to assess the consistency and discrepancies between the output mask and the ground truth. F1 score is a comprehensive evaluation metric that quantitatively analyzes each pixel, represented as the harmonic average value of precision and recall \cite{shengao2}.
\begin{equation}
    F1=2\times\frac{ Precision\times Recall}{Precision+Recall} ,
\end{equation}
where
\begin{equation}
    Precision=\frac{TP}{TP+FP},
\end{equation}
\begin{equation}
    Recall=\frac{TP}{TP+FN} .
\end{equation}

Here $TP$ represents the number of truly positive samples correctly classified as positive, $FP$ represents the number of truly negative samples incorrectly classified as positive, and $FN$ represents the number of actually positive samples incorrectly classified as negative.
\paragraph{Metrics in Land-Cover Classification}
For land-cover classification tasks, we adopt the mean Average Precision (mAP) to measure the quality of classification results. AP represents the average precision for a single class label, and its value corresponds to the area under the Precision-Recall curve. And mAP stands for the average of the individual AP values calculated for all $N$ categories:
\begin{equation}
    mAP=\frac{AP}{N}=\frac{\int_{0}^{1}p(r)dr }{N}  ,
\end{equation}
where $p$ refers to Precision and $r$ denotes Recall \cite{superyolo}.
\begin{figure*}[tbp]
    \centering
    \includegraphics[width=1\textwidth]{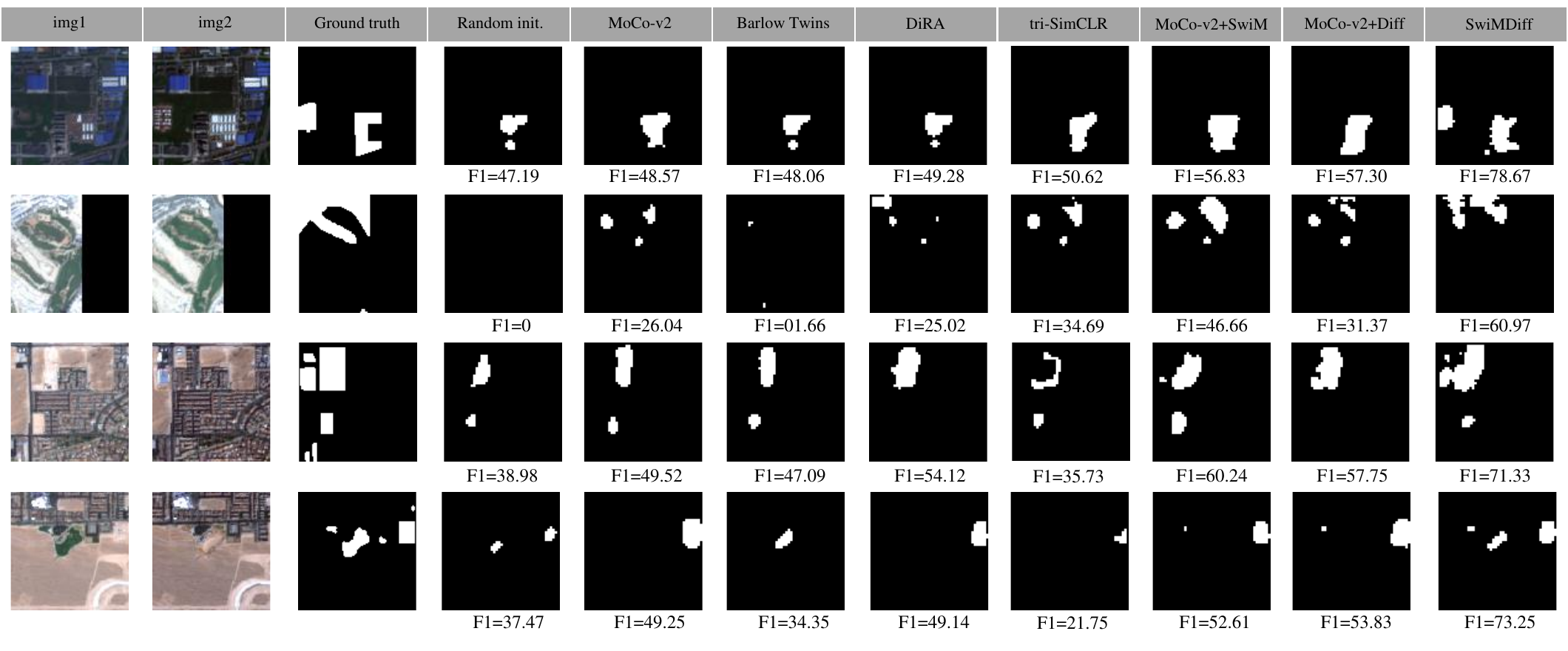}
    \caption{Comparison and ablation of visualisation results on the Onera Satellite change detection task.}
    \label{oscd}
\end{figure*}
\subsection{Change Detection on Onera Satellite}
This task focuses on detecting changes in image pairs captured from the same geographic location but at different times. We evaluate the performance using the F1 score and utilize the Onera Satellite Change Detection (OSCD) \cite{oscd} dataset. This dataset comprises 24 pairs of multispectral images across 13 spectral bands of Sentinel-2, with resolutions of 10 m, 20 m, and 60 m. For each pair of multi-temporal images, there exists a corresponding change detection label, with white pixels indicating changes and black pixels representing no change. Our experiments are confined to the RGB bands, and the dataset is divided for training and validation following the methodology established by \emph{et al.} \cite{oscd}.

\subsubsection{Implementation Details} 
\label{Section 4.2.2}
The network setup for this task, as shown in Fig. \ref{change_kuangtu}, involves processing each pair of images into two sets of features using the pre-trained ResNet-18 \cite{resnet} backbone. The absolute difference between these feature sets is then calculated and input into a U-Net \cite{unet} decoder to create change detection masks. For this task, we keep the ResNet-18 backbone static and focus on training the U-Net's remaining components for 100 epochs, using a batch size of 32. In line with the suggestions of Manas \emph{et al.} \cite{seco}, our approach includes image augmentation through random horizontal flips and 90° rotations. We use an Adam optimizer \cite{adam} with a weight decay of 1e-4 and an initial learning rate of 1e-3.

\begin{figure}[tbp]
    \centering
    \includegraphics[width=0.48\textwidth]{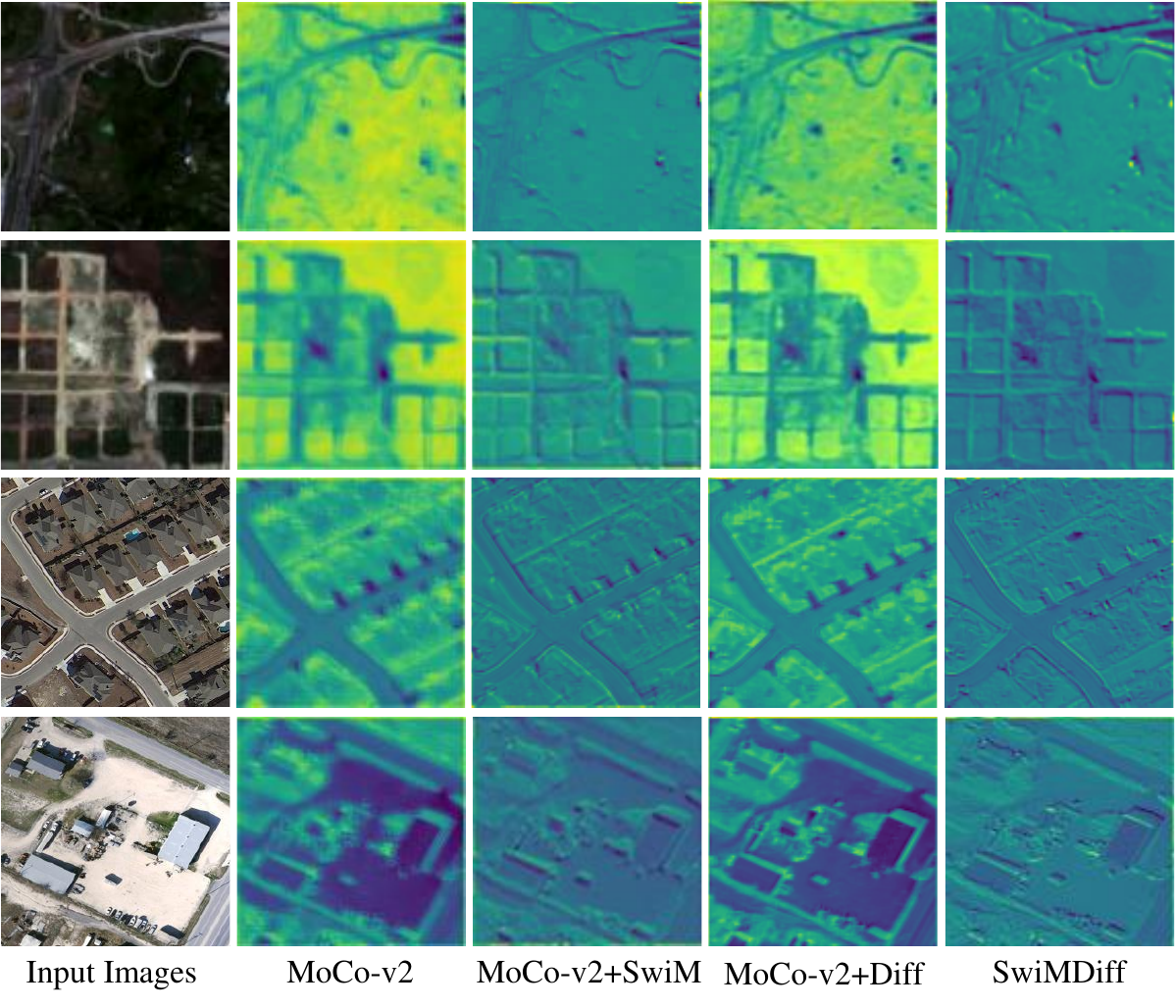}
    \caption{Qualitative results of our experiments. The top two rows are from OSCD and the bottom two rows are from LEVIR-CD. It can be seen that the extracted features after pre-training of different methods are all enhanced on the baseline.
    }
    \label{tezheng}
\end{figure}
\begin{table}[tpb]
	\small
	\centering
	\setlength{\tabcolsep}{1.8mm}{
		\caption{the Comparison Results of Precision, Recall and F1(\%) on the Change Detection of OSCD Validation Set}
		\label{tbl:oscd}
        \renewcommand\arraystretch{1.2}
		\begin{tabular}{c|ccc}
			\toprule[1.2pt]
			\textbf{Method}  & \textbf{Precision(\%)}  & \textbf{Recall(\%)} & \textbf{F1(\%)} $\uparrow$ \\
			\midrule
			  Random init. & 69.8   & 23.5  & 34.1     \\
                MoCo-v2 & 54.4 & 40.7  &45.6  \\
			Barlow Twins  & 59.8  & 35.7   & 44.2   \\
			DiRA & 63.5    & 37.9          & 46.6   \\
            tri-SimCLR & 62.3    & 37.6          & 46.2   \\
                SwiMDiff(ours) &63.6 & 40.9  & \textbf{49.6} \\
			\bottomrule[1.2pt]
	\end{tabular}}
\vspace{-0.1in}
\end{table}
\subsubsection{Results Discussion} 
\paragraph{Comparison Results}
\begin{figure*}[htbp]
    \centering
    \includegraphics[width=1\textwidth]{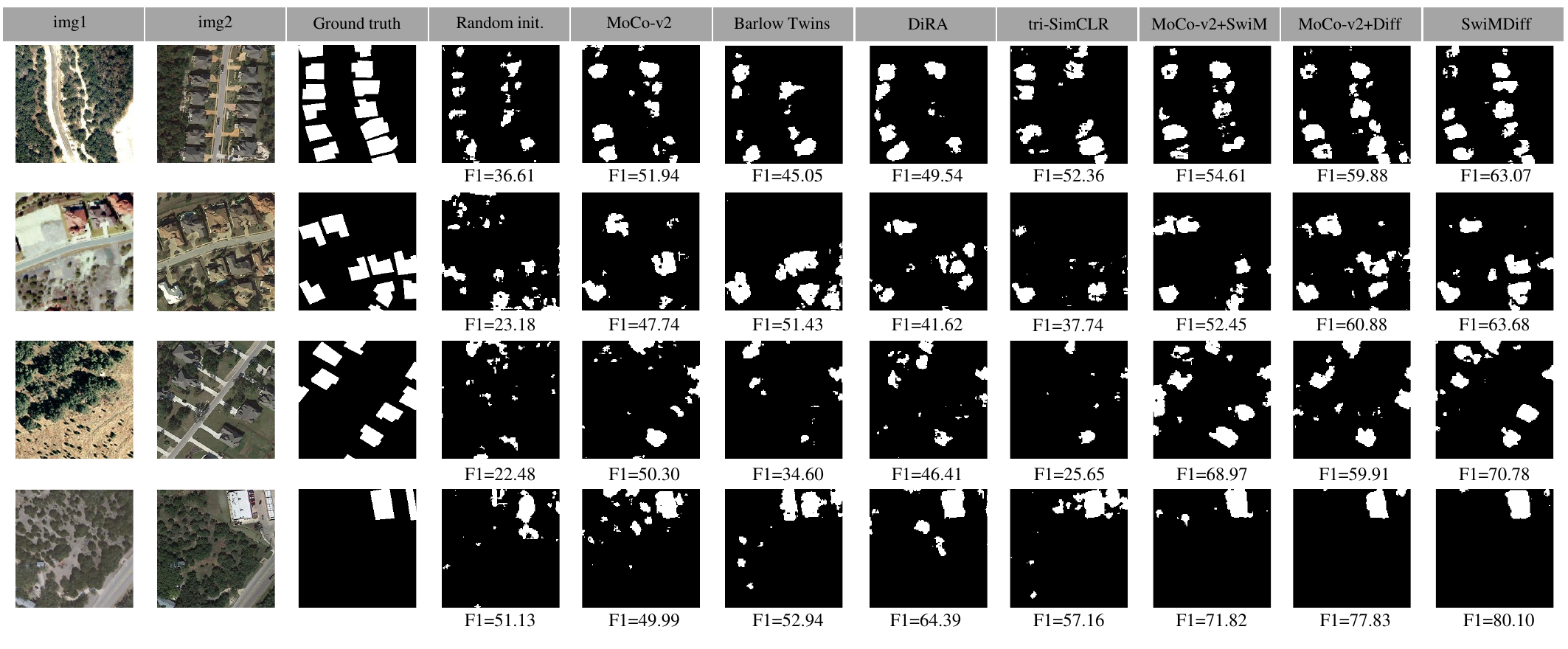}
    \caption{Comparison and ablation of visualisation results on the LEVIR-CD change detection task.}
    \label{levir}
\end{figure*}

\begin{table*}[htpb]
	\small
	\centering
	\setlength{\tabcolsep}{1.8mm}{
		\caption{the Ablation Results of Precision, Recall and F1(\%) on the Change Detection of OSCD and LEVIR-CD Validation Sets}
		\label{cdab}
        \renewcommand\arraystretch{1.3}
        \tabcolsep=0.2cm
        \begin{tabular}{cc|ccc|ccc} 
        
        \toprule[1.2pt]

\multicolumn{2}{c|}{\textbf{Components} }                                                       & \multicolumn{3}{c|}{\textbf{OSCD}} & \multicolumn{3}{c}{\textbf{LEVIR-CD}}  \\ 
    \midrule
  \multicolumn{1}{c}{SwiM} & \multicolumn{1}{c|}{Diff} & Precision(\%)& Recall(\%) & F1(\%)$\uparrow$              & Precision(\%)& Recall(\%) & F1(\%)$\uparrow$                     \\ 
\midrule
                             &                            &  54.4 &  40.7 &  45.6          & 81.7 & 76.8 &  79.1          \\
                             $\checkmark$    &                       
                            & 50.2 &45.1 & 46.7       &82.7 &77.5&80.0                     \\
                                            &   $\checkmark$          
                        & 61.8& 41.3&48.6              & 82.9&77.6 &80.1                     \\
                       $\checkmark$     &    $\checkmark$           
                    &63.6 &40.9& \textbf{49.6}                &83.6 &78.3 &\textbf{80.9}                     \\
            \bottomrule[1.2pt]
        \end{tabular}}
        \vspace{-0.1in}
\end{table*}

As presented in Table \ref{tbl:oscd}, we evaluate SwiMDiff against various methods, including random initialization, MoCo-v2 \cite{mocov2} (our baseline), Barlow Twins \cite{cl4barlow}, DiRA \cite{dira}, and tri-SimCLR \cite{tri}. SwiMDiff achieves the highest F1 scores, demonstrating its superior capability in detecting changes. Notably, SwiMDiff exceeds the baseline's performance by a significant 4.0\% margin in F1 score. Qualitative analyses, showcased in the top two rows of Fig. \ref{tezheng}, reveal that SwiMDiff extracts more precise image details and sharper object boundaries compared to the baseline. These improvements underscore our method's enhanced ability to extract shallow pixel-level features and detail information of the image. The detection masks generated by SwiMDiff, as exhibited in Fig. \ref{oscd}, more closely match the ground truth, covering a larger number of changed pixels, thereby indicating superior quality over other methods.


\paragraph{Ablation Results}
The ablation study results, presented in Table \ref{cdab} and Fig. \ref{oscd}, demonstrate the enhancements achieved by adding single modules to the baseline. The inclusion of scene-wide matching (MoCo-v2+SwiM) improves the F1 score by 1.1\%, while integrating the diffusion model (MoCo-v2+Diff) leads to a 3.0\% increase. As evident in Fig. \ref{tezheng}, features refined through the diffusion auxiliary task are markedly clearer and more defined compared to the baseline. Furthermore, the MoCo-v2+SwiM module also shows an improvement in the extraction of detailed features. This aligns with our hypothesis that incorporating a diffusion constraint deepens the understanding of detail and pixel-level information in images. SwiMDiff, by combining these two modules, significantly enhances the original self-supervised method, resulting in more nuanced and fine-grained change detection.


\subsection{Change Detection on LEVIR-CD}
LEVIR-CD \cite{levir} is a comprehensive remote sensing change detection dataset that includes bitemporal image pairs from 20 diverse regions in Texas, USA, spanning 5-14 years. This dataset encapsulates a wide array of buildings, such as villas residences, tall apartments, small garages and large warehouses. It comprises 637 very high-resolution image pairs from Google Earth\cite{google}, each with dimensions of $1024\times 1024$ pixels, meticulously inspected to ensure quality.


\subsubsection{Implementation Details} 
For this task, we implement the same architectural model as used in the Change Detection on Onera Satellite task (Section \ref{Section 4.2.2}). Our approach utilizes a pre-trained, frozen ResNet-18 \cite{resnet} to extract features. The training is conducted over 100 epochs with a batch size of 32, using an Adam optimizer \cite{adam} with an initial learning rate of 1e-3 and a weight decay of 1e-4. We follow the data partitioning strategy suggested by Chen \emph{et al.} \cite{levir} for training and validation, segmenting the images into $256\times 256$ non-overlapping tiles and disregarding completely black labels.


\begin{table}[tpb]
	\small
	\centering
	\setlength{\tabcolsep}{1.8mm}{
		\caption{the Comparison and Ablation Results of Precision, Recall and F1(\%) on the Change Detection of LEVIR-CD Validation Set}
		\label{tbl:levir}
        \renewcommand\arraystretch{1.2}
		\begin{tabular}{c|ccc}
			\toprule[1.2pt]
			\textbf{Method}  & \textbf{Precision(\%)}  & \textbf{Recall(\%)} & \textbf{F1(\%)} $\uparrow$ \\
			\midrule
			  Random init. & 67.3   & 44.1  & 53.1     \\
                MoCo-v2 & 81.7 & 76.8  &79.1  \\
			Barlow Twins  & 80.1  & 75.3   & 77.5   \\
			DiRA & 83.5    & 75.0          & 79.0   \\
            tri-SimCLR & 83.2    & 75.9         & 79.5   \\
                SwiMDiff(ours) &83.6 & 78.3  & \textbf{80.9} \\
			\bottomrule[1.2pt]
	\end{tabular}}
\vspace{-0.1in}
\end{table}
\subsubsection{Results Discussion} 
\paragraph{Comparison Results}
As indicated in Table \ref{tbl:levir}, SwiMDiff outperforms other methods in precision, recall, and F1 score, demonstrating a notable 1.8\% improvement in F1 score over the baseline. This underlines its superior performance in change detection. In Fig. \ref{tezheng}, the bottom two rows clearly show that SwiMDiff extracts more detailed and sharper features compared to the baseline, which are crucial for accurate image analysis. Fig. \ref{levir} displays the detection masks generated by various methods on the LEVIR-CD validation set, where SwiMDiff's masks show higher quality in more complex scenarios, detecting more changes and reducing large-scale misses. It highlights SwiMDiff's advancement in change detection performance, both qualitatively and quantitatively, across diverse datasets and challenging scenarios.


\paragraph{Ablation Results}
The ablation study, detailed in Table \ref{cdab} and Fig. \ref{levir}, reveals that MoCo-v2+SwiM and MoCo-v2+Diff respectively improve F1 scores by 0.9\% and 1.0\% compared to the baseline. This demonstrates the effectiveness of each module when applied independently. Notably, the scene-wide matching module also shows improvements in extracting detailed features and accomplishing pixel-level tasks. SwiMDiff, combining these two modules, further refines feature extraction, enhancing the overall performance in change detection.

\begin{figure}[tbp]
    \centering
    \includegraphics[width=0.48\textwidth]{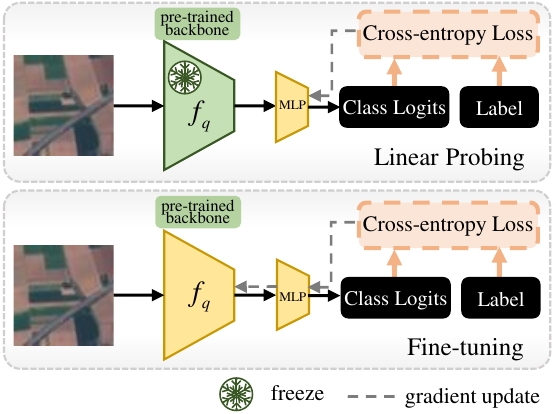}
    \caption{The network architecture for land-cover classification task. We load pre-trained weights as the initial weights for the encoder$f_{q}$. In the case of linear probing, the encoder's weights are frozen, while only the MLP layer used for classification is updated with gradients. In the case of fine-tuning, both the encoder and the classifier are updated with gradients.}
    \label{fenlei_kuangtu}
\end{figure}

\subsection{Land-Cover Classification on BigEarthNet}
BigEarthNet \cite{bigearthnet}, a comprehensive dataset, consists of 590,326 distinct multispectral images from Sentinel-2 \cite{s2}, sourced from 125 tiles covering 10 European countries. Designed for multi-label land-cover classification, it encompasses 19 diverse categories, with each image annotated with multiple classes. Covering $1.2\times 1.2$ km per image at resolutions of 10 m, 20 m, and 60 m per pixel, we selectively use 10\% of the dataset, excluding images fully obscured by snow, clouds, or shadows, for training and validation.

\begin{table}[tpb]
	\small
	\renewcommand{\arraystretch}{1.2}
	\centering
	\setlength{\tabcolsep}{1.8mm}{
		\caption{the Comparison Results of mAP on the Land-cover Classification of BigEarthnet Validation Set}
		\label{bigearth}
		\begin{tabular}{c|cc}
			\toprule[1.2pt]
			\textbf{Method}  & \textbf{Linear Probing(\%)} $\uparrow$ & \textbf{Fine-tuning(\%)} $\uparrow$ \\
			\midrule
			  Random init. & 43.6   & 69.4   \\
                MoCo-v2 & 68.6 & 80.4  \\
			Barlow Twins  & 61.1  & 79.2    \\
			DiRA & 69.7    & 80.7           \\
            tri-SimCLR & 67.5    & 79.6\\ 
                SwiMDiff(ours) &\textbf{69.9} & \textbf{81.1}   \\
			\bottomrule[1.2pt]
	\end{tabular}}
\vspace{-0.1in}
\end{table}
\begin{table}
    \centering
    \setlength{\tabcolsep}{1.8mm}{
		\caption{the Ablation Experiments on Different Weight Coefficients During the Pre-training Phase}
		\label{tbl:lamda}
        \renewcommand\arraystretch{1.2}
        \begin{tabular}{c|ccccc} 
        \toprule[1.2pt]
        $\sfrac{\lambda_C}{\lambda_D}$        & $\sfrac{1}{14}$ & $\sfrac{1}{12}$   & $\sfrac{1}{10}$   & $\sfrac{1}{8}$   & $\sfrac{1}{6}$    \\ 
        \midrule
        Accuracy(\%) & 69.3 & 69.8 & \textbf{69.9} & 69.8 & 69.5  \\
        \bottomrule[1.2pt]
        \end{tabular}
        \vspace{-0.1in}}
\end{table}

\subsubsection{Implementation Details} 
Our classification task utilize a pre-trained ResNet-18 \cite{resnet} as the backbone, coupled with an additional MLP layer for classifying high-dimensional features. The network is trained for 100 epochs with a batch size of 256. Specifically, as demonstrated in the Fig. \ref{fenlei_kuangtu}, for the linear probing approach, we freeze the backbone and employ an Adam optimizer \cite{adam} with an initial learning rate set to 1e-3, focusing on iteratively optimizing the classifier. For the fine-tuning approach, we adjust the entire network including the backbone and classifier with an initial learning rate of 1e-5. The learning rate is reduced by 10 at 60\% and 80\% of the total epochs.



\subsubsection{Results Discussion}
\paragraph{Comparison Results}
As depicted in Table \ref{bigearth}, SwiMDiff significantly outperforms other methods in pre-training accuracy, particularly evident with a 1.3\% increase in linear probing performance over the baseline. Notably, when the entire network is fine-tuned, performance differences among methods are less pronounced. In the fine-tuning phase, SwiMDiff exhibits an increment of just 0.7\% observed. This underscores SwiMDiff's enhanced ability in extracting global discriminative information and high-level semantic features.


\paragraph{Ablation Results}
Table \ref{tbl:abcla} details the impact of individual modules. Compared to the baseline, the accuracy of linear probing with MoCo-v2+SwiM increases by 1.1\%, while the fine-tuning result makes a mere 0.3\% improvement. For MoCo-v2+Diff, there is a 1.2\% enhancement in linear probing accuracy, and a 0.5\% increase for fine-tuned outcomes. Both modules individually contribute to the elevation of detection precision. The scene-wide matching technique and the auxiliary task of the diffusion model both serve to bolster the model’s ability to learn global semantic features.
\begin{figure*}[htbp]
    \centering
   \includegraphics[width=1\textwidth]{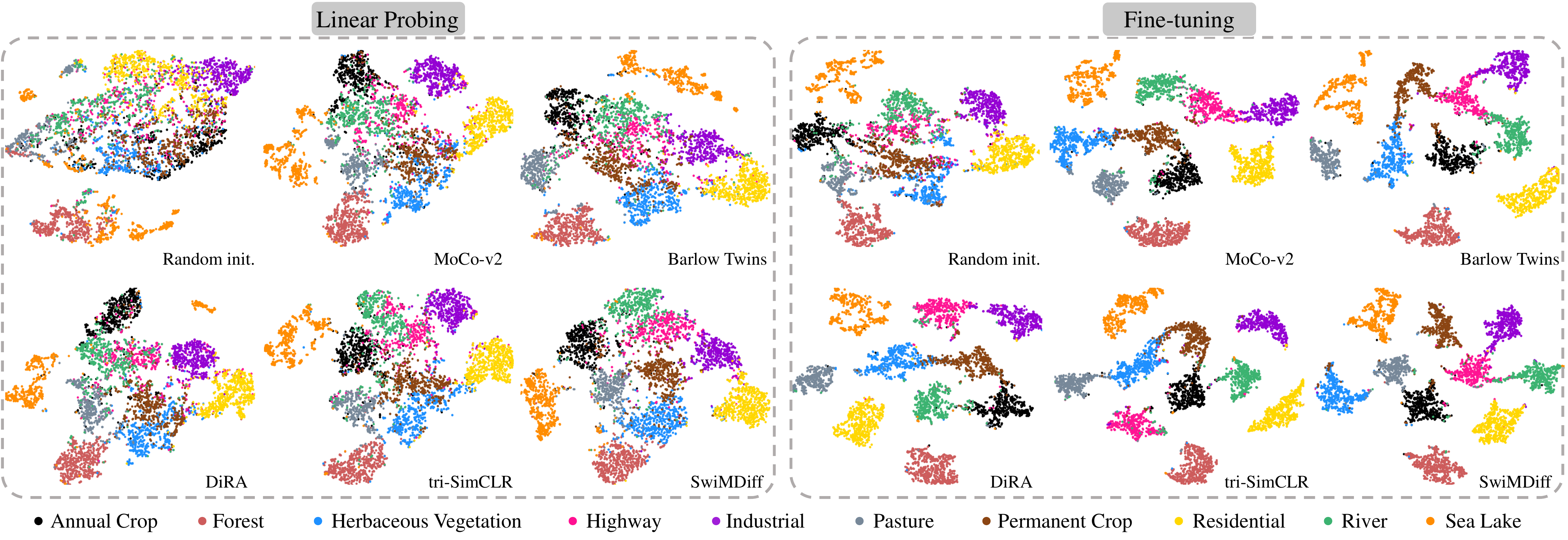}
    \caption{The t-SNE visualization of learned representations on the validation set of EuroSAT.}
    \label{julei}
\end{figure*}
\begin{table*}[tpb]
	\small
	\centering
	\setlength{\tabcolsep}{1.8mm}{
		\caption{the Ablation Results of mAP on the Land-cover Classification of BigEarthnet and EuroSAT Validation Sets}
		\label{tbl:abcla}
        \renewcommand\arraystretch{1.3}
        \tabcolsep=0.2cm
        \begin{tabular}{cc|cc|cc} 
        
        \toprule[1.2pt]

\multicolumn{2}{c|}{\textbf{Components} }                                                       & \multicolumn{2}{c|}{\textbf{BigEarthNet}} & \multicolumn{2}{c}{\textbf{EuroSAT}}  \\ 
    \midrule
  \multicolumn{1}{c}{SwiM} & \multicolumn{1}{c|}{Diff} & Linear Probing(\%)$\uparrow$   & Fine-tuning(\%)$\uparrow$    & Linear Probing(\%)$\uparrow$   & Fine-tuning(\%)$\uparrow$      \\ 
\midrule
                             &          &  68.6 &  80.4          
                             & 86.6 & 94.8           \\
                             $\checkmark$    &                       
                            & 69.7 &80.7       &88.5 &95.7     \\
                                            &   $\checkmark$          
                        & 69.8&80.9   & 88.3 &95.4                 \\
                       $\checkmark$     &    $\checkmark$           
     & \textbf{69.9}&\textbf{81.1} &\textbf{89.1} &\textbf{96.1}     \\
            \bottomrule[1.2pt]
        \end{tabular}}
        \vspace{-0.1in}
\end{table*}

\paragraph{Ablation on Weight Factors}
To achieve optimal training results for the SwiMDiff, we design ablation experiments for the values of $\lambda _{C}$ and $\lambda _{D}$. We maintain the weight of CL loss at $1$, while varying the weight of the diffusion loss during the training process. We test and evaluate pre-trained models with different weight ratios on the classification task of BigEarthNet with linear probing, resulting in Table \ref{tbl:lamda}. We find that the weight coefficient ratio is very crucial during the pre-training phase. As shown in Table \ref{tbl:lamda}, the best integration effect of CL and diffusion model occurs when the weight coefficients are $1$ and $10$, respectively. However, when the weight coefficient ratio is too large or too small, it is not conducive to the integration of them.

\subsection{Land-Cover Classification on EuroSAT}
EuroSAT \cite{EuroSAT} is a single-label land-cover classification dataset, featuring images with dimensions of $64\times 64$ pixels. Originating from the Sentinel-2 satellite, it encompasses imagery across 13 spectral bands. Within EuroSAT, there are 10 distinct classes, each containing between 2,000 to 3,000 images, summing up to a total of 27,000 images. Noteworthy, the dataset is curated to exclude images with high cloud coverage but does not undergo atmospheric correction. We partition the data into training and validation sets following the method proposed in \cite{neumann2019domain}.
\subsubsection{Implementation Details} 

We attach a linear classification head after the pre-trained backbone and train it for 100 epochs with a batch size of 256. The training process, as shown in Fig. \ref{fenlei_kuangtu}, is divided into linear probing and fine-tuning phases, each with respective initial learning rates of 1e-3 and 1e-5, reduced by 0.1 at the 60\% and 80\% epochs.
\begin{table}[tpb]
	\small
	\renewcommand{\arraystretch}{1.2}
	\centering
	\setlength{\tabcolsep}{1.8mm}{
		\caption{the Comparison Results of mAP on the Land-cover Classification of EuroSAT Validation Set}
		\label{euro}
		\begin{tabular}{c|cc}
			\toprule[1.2pt]
			\textbf{Method}  & \textbf{Linear Probing(\%)} $\uparrow$ & \textbf{Fine-tuning(\%)} $\uparrow$ \\
			\midrule
			  Random init. & 69.5   & 85.8   \\
                MoCo-v2 & 86.6 & 94.8  \\
			Barlow Twins  & 86.8  & 95.0    \\
			DiRA & 87.5    & 95.2           \\
            tri-SimCLR & 88.6    &  95.9         \\
                SwiMDiff(ours) &\textbf{89.1} & \textbf{96.1}   \\
			\bottomrule[1.2pt]
	\end{tabular}}
\vspace{-0.1in}
\end{table}

\subsubsection{Results Discussion}
\begin{figure}[htbp]
    \centering
   \includegraphics[width=0.48\textwidth]{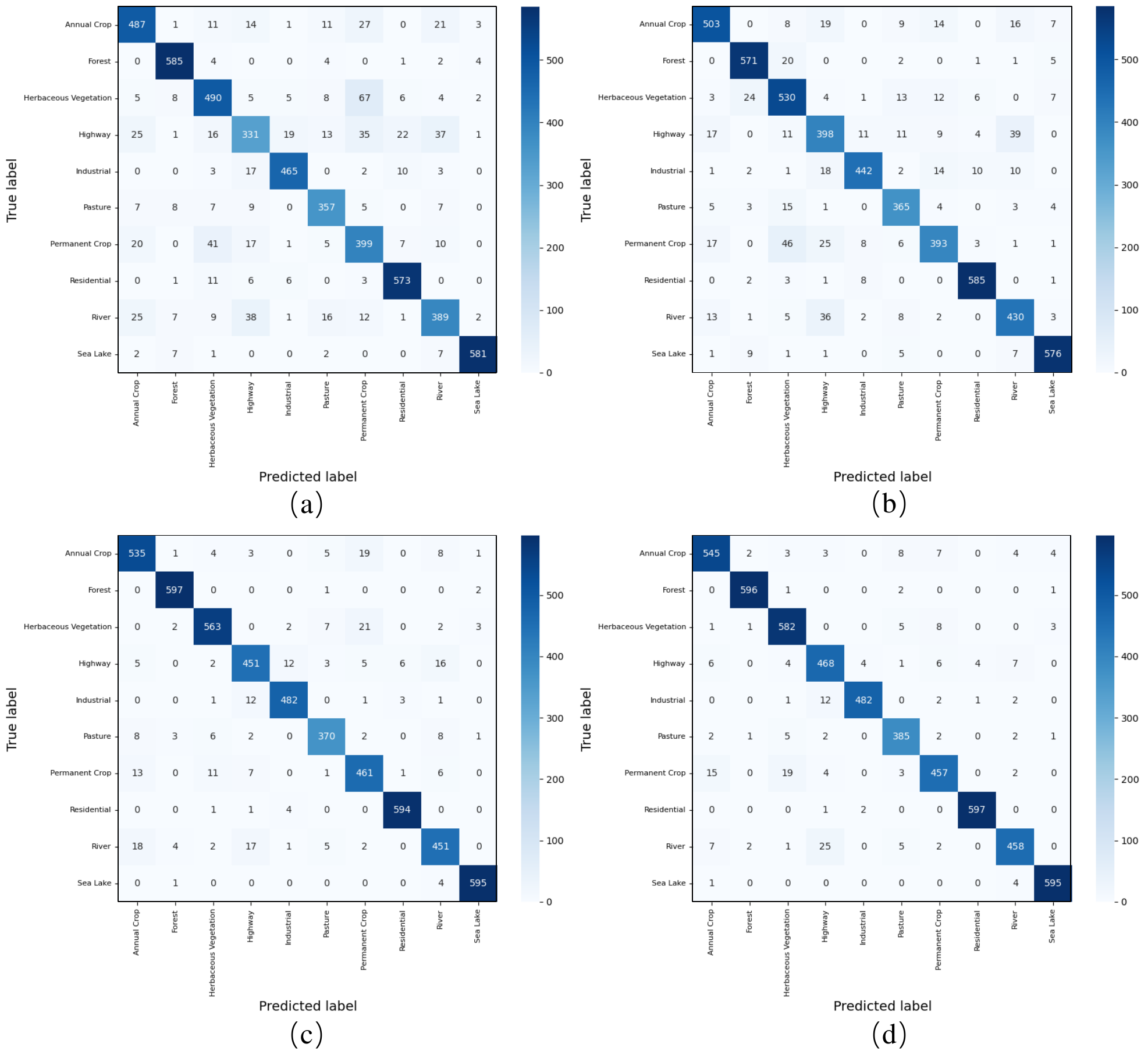}
    \caption{Confusion matrices on the EuroSAT validation set. (a): Linear probing confusion matrix of MoCo-v2. (b): Linear probing confusion matrix of SwiMDiff. (c): Fine-tuning confusion matrix of MoCo-v2. (d): Fine-tuning confusion matrix of SwiMDiff.}
    \label{hunxiao}
\end{figure}

\paragraph{Comparison Results}
As shown in Table \ref{euro}, SwiMDiff achieved the best classification results on the Eurosat dataset, improving by 2.5\% over MoCo-v2 in linear probing and by 1.3\% in fine-tuning. Fig. \ref{julei} and Fig. \ref{hunxiao} respectively display the t-SNE clustering visualization and the confusion matrix for the classification results on the EuroSAT validation set. In Fig. \ref{julei}, compared to MoCo-v2, SwiMDiff demonstrates a more distinct and effective ability to cluster the same categories and separate different ones.  For instance, in linear probing, `Herbaceous Vegetation', `Pasture', and `River' are each more tightly clustered, while in fine-tuning, the boundaries distinguishing `Highway' from `River' and `Industrial' are clearer. The confusion matrices in Fig. \ref{hunxiao} present specific classification results, also demonstrating the advanced nature and superiority of our proposed SwiMDiff.
 

\paragraph{Ablation Results}
Table \ref{tbl:abcla} also displays the classification ablation results of each module on the EuroSAT dataset. For linear probing, MoCo-v2+SwiM and MoCo-v2+Diff respectively improve by 1.9\% and 1.7\% over the baseline, while for fine-tuning, they improve by 0.9\% and 0.6\% respectively. These results indicate the effectiveness of these modules in extracting discriminative information and semantic details, underscoring our method's versatility and potential applicability to various remote sensing datasets.


\section{Conclusion and Future Work}
\label{sec:Conclusion}

To enhance the efficacy of CL in remote sensing, we introduce SwiMDiff, a novel self-supervised pretraining framework. Initially, our approach incorporates a scene-wide matching strategy into CL. This strategy intoroduces intra-class similarityby treating images from the same scene as false negatives, thereby effectively addressing sample confusion and enhancing representation learning. Additionally, SwiMDiff integrates the diffusion model with CL. It utilizes pixel-level diffusion constraints to amplify the encoder's ability in detail extraction, particularly emphasizing fine-grained details in RSI. These two aspects complement each other, collectively strengthening the encoder's ability to extract both global and local features from RSI. SwiMDiff contributes a richer and more transferable representation for remote sensing, presenting a new self-supervised solution for unlabeled RSIs.

Under consistent conditions with regards to pretraining datasets and backbone networks, we compare SwiMDiff with state-of-the-art self-supervised methods. Experimental results demonstrate the superiority of SwiMDiff in change detection on OSCD and LEVIR-CD, as well as land-cover classification on BigEarthNet and EuroSAT.

In the continuation of our work, we will focus on compressing and accelerating the network framework integrated with the diffusion model, aiming to reduce the computational resources required for image self-supervised representation learning to make it more practical and efficient.





{
	\bibliographystyle{IEEEtran}
	\bibliography{reference}

\begin{thebibliography}{10}
\providecommand{\url}[1]{#1}
\csname url@samestyle\endcsname
\providecommand{\newblock}{\relax}
\providecommand{\bibinfo}[2]{#2}
\providecommand{\BIBentrySTDinterwordspacing}{\spaceskip=0pt\relax}
\providecommand{\BIBentryALTinterwordstretchfactor}{4}
\providecommand{\BIBentryALTinterwordspacing}{\spaceskip=\fontdimen2\font plus
\BIBentryALTinterwordstretchfactor\fontdimen3\font minus
  \fontdimen4\font\relax}
\providecommand{\BIBforeignlanguage}[2]{{%
\expandafter\ifx\csname l@#1\endcsname\relax
\typeout{** WARNING: IEEEtran.bst: No hyphenation pattern has been}%
\typeout{** loaded for the language `#1'. Using the pattern for}%
\typeout{** the default language instead.}%
\else
\language=\csname l@#1\endcsname
\fi
#2}}
\providecommand{\BIBdecl}{\relax}
\BIBdecl

\bibitem{gao3}
Z.~Chen, D.~Hong, and H.~Gao, ``Grid network: Feature extraction in anisotropic
  perspective for hyperspectral image classification,'' \emph{IEEE Geosci.
  Remote Sens. Lett.}, vol.~20, pp. 1--5, 2023.

\bibitem{gao4}
\BIBentryALTinterwordspacing
Z.~Chen, G.~Wu, H.~Gao, Y.~Ding, D.~Hong, and B.~Zhang, ``Local aggregation and
  global attention network for hyperspectral image classification with
  spectral-induced aligned superpixel segmentation,'' \emph{Expert Syst.
  Appl.}, vol. 232, p. 120828, 2023. [Online]. Available:
  \url{https://www.sciencedirect.com/science/article/pii/S0957417423013301}
\BIBentrySTDinterwordspacing

\bibitem{gao1}
\BIBentryALTinterwordspacing
Z.~Chen, Y.~Wang, H.~Gao, Y.~Ding, Q.~Zhong, D.~Hong, and B.~Zhang, ``Temporal
  difference-guided network for hyperspectral image change detection,''
  \emph{Int. J. Remote Sens.}, vol.~44, no.~19, pp. 6033--6059, 2023. [Online].
  Available: \url{https://doi.org/10.1080/01431161.2023.2258563}
\BIBentrySTDinterwordspacing

\bibitem{shengao1}
\BIBentryALTinterwordspacing
Q.~Zhu, X.~Guo, Z.~Li, and D.~Li, ``A review of multi-class change detection
  for satellite remote sensing imagery,'' \emph{Geo-Spat. Inf. Sci.}, vol.~0,
  no.~0, pp. 1--15, 2022. [Online]. Available:
  \url{https://doi.org/10.1080/10095020.2022.2128902}
\BIBentrySTDinterwordspacing

\bibitem{shengao2}
\BIBentryALTinterwordspacing
Q.~Zhu, X.~Guo, W.~Deng, S.~Shi, Q.~Guan, Y.~Zhong, L.~Zhang, and D.~Li,
  ``Land-use/land-cover change detection based on a siamese global learning
  framework for high spatial resolution remote sensing imagery,''
  \emph{ISPRS-J. Photogramm. Remote Sens.}, vol. 184, pp. 63--78, 2022.
  [Online]. Available:
  \url{https://www.sciencedirect.com/science/article/pii/S0924271621003270}
\BIBentrySTDinterwordspacing

\bibitem{gao2}
Z.~Chen, Z.~Lu, H.~Gao, Y.~Zhang, J.~Zhao, D.~Hong, and B.~Zhang, ``Global to
  local: A hierarchical detection algorithm for hyperspectral image target
  detection,'' \emph{IEEE Trans. on Geosci. and Remote Sens.}, vol.~60, pp.
  1--15, 2022.

\bibitem{ghost}
J.~Zhang, J.~Lei, W.~Xie, Y.~Li, G.~Yang, and X.~Jia, ``Guided hybrid
  quantization for object detection in remote sensing imagery via one-to-one
  self-teaching,'' \emph{IEEE Trans. on Geosci. and Remote Sens.}, 2023.

\bibitem{superyolo}
J.~Zhang, J.~Lei, W.~Xie, Z.~Fang, Y.~Li, and Q.~Du, ``Superyolo: Super
  resolution assisted object detection in multimodal remote sensing imagery,''
  \emph{IEEE Trans. on Geosci. and Remote Sens.}, vol.~61, pp. 1--15, 2023.

\bibitem{sader1990remote}
S.~Sader, T.~Stone, and A.~Joyce, ``Remote sensing of tropical forests- an
  overview of research and applications using non-photographic sensors,''
  \emph{Photogramm. Eng. Remote Sens.}, vol.~56, no.~10, pp. 1343--1351, 1990.

\bibitem{schumann2018assisting}
G.~J. Schumann, G.~R. Brakenridge, A.~J. Kettner, R.~Kashif, and E.~Niebuhr,
  ``Assisting flood disaster response with earth observation data and products:
  A critical assessment,'' \emph{Remote Sens.}, vol.~10, no.~8, p. 1230, 2018.

\bibitem{mulla2013twenty}
D.~J. Mulla, ``Twenty five years of remote sensing in precision agriculture:
  Key advances and remaining knowledge gaps,'' \emph{Biosyst. Eng.}, vol. 114,
  no.~4, pp. 358--371, 2013.

\bibitem{filipponi2019exploitation}
F.~Filipponi, ``Exploitation of sentinel-2 time series to map burned areas at
  the national level: A case study on the 2017 italy wildfires,'' \emph{Remote
  Sens.}, vol.~11, no.~6, p. 622, 2019.

\bibitem{berg2022self}
P.~Berg, M.-T. Pham, and N.~Courty, ``Self-supervised learning for scene
  classification in remote sensing: Current state of the art and
  perspectives,'' \emph{Remote Sens.}, vol.~14, no.~16, p. 3995, 2022.

\bibitem{wang2022self}
\BIBentryALTinterwordspacing
Y.~Wang, C.~M. Albrecht, N.~A.~A. Braham, L.~Mou, and X.~X. Zhu,
  ``Self-supervised learning in remote sensing: A review,'' \emph{arXiv}, 2022.
  [Online]. Available: \url{https://arxiv.org/abs/2206.13188}
\BIBentrySTDinterwordspacing

\bibitem{peng1}
Q.~Liu, J.~Peng, N.~Chen, W.~Sun, Y.~Ning, and Q.~Du, ``Category-specific
  prototype self-refinement contrastive learning for few-shot hyperspectral
  image classification,'' \emph{IEEE Trans. on Geosci. and Remote Sens.},
  vol.~61, pp. 1--16, 2023.

\bibitem{peng2}
Q.~Liu, J.~Peng, Y.~Ning, N.~Chen, W.~Sun, Q.~Du, and Y.~Zhou, ``Refined
  prototypical contrastive learning for few-shot hyperspectral image
  classification,'' \emph{IEEE Trans. on Geosci. and Remote Sens.}, vol.~61,
  pp. 1--14, 2023.

\bibitem{peng3}
Y.~Ning, J.~Peng, Q.~Liu, Y.~Huang, W.~Sun, and Q.~Du, ``Contrastive learning
  based on category matching for domain adaptation in hyperspectral image
  classification,'' \emph{IEEE Trans. on Geosci. and Remote Sens.}, vol.~61,
  pp. 1--14, 2023.

\bibitem{cookbook}
\BIBentryALTinterwordspacing
R.~Balestriero, M.~Ibrahim, V.~Sobal, A.~Morcos, S.~Shekhar, T.~Goldstein,
  F.~Bordes, A.~Bardes, G.~Mialon, Y.~Tian \emph{et~al.}, ``A cookbook of
  self-supervised learning,'' \emph{arXiv}, 2023. [Online]. Available:
  \url{https://arxiv.org/abs/2304.12210}
\BIBentrySTDinterwordspacing

\bibitem{jigsaw}
M.~Noroozi and P.~Favaro, ``Unsupervised learning of visual representations by
  solving jigsaw puzzles,'' in \emph{Proc. Eur. Conf. Comput. Vis.
  (ECCV)}.\hskip 1em plus 0.5em minus 0.4em\relax Springer, 2016, pp. 69--84.

\bibitem{patch}
C.~Doersch, A.~Gupta, and A.~A. Efros, ``Unsupervised visual representation
  learning by context prediction,'' in \emph{Proc. IEEE/CVF Int. Conf. Comput.
  Vis. (ICCV)}, 2015, pp. 1422--1430.

\bibitem{inpainting}
R.~Zhang, P.~Isola, and A.~A. Efros, ``Colorful image colorization,'' in
  \emph{Proc. Eur. Conf. Comput. Vis. (ECCV)}.\hskip 1em plus 0.5em minus
  0.4em\relax Springer, 2016, pp. 649--666.

\bibitem{cl1}
M.~Caron, I.~Misra, J.~Mairal, P.~Goyal, P.~Bojanowski, and A.~Joulin,
  ``Unsupervised learning of visual features by contrasting cluster
  assignments,'' \emph{Proc. Adv. Neural Inf. Process. Syst.}, vol.~33, pp.
  9912--9924, 2020.

\bibitem{moco}
K.~He, H.~Fan, Y.~Wu, S.~Xie, and R.~Girshick, ``Momentum contrast for
  unsupervised visual representation learning,'' in \emph{Proc. IEEE/CVF Int.
  Conf. Comput. Vis. (ICCV)}, 2020, pp. 9729--9738.

\bibitem{mocov2}
\BIBentryALTinterwordspacing
X.~Chen, H.~Fan, R.~Girshick, and K.~He, ``Improved baselines with momentum
  contrastive learning,'' \emph{arXiv}, 2020. [Online]. Available:
  \url{https://arxiv.org/abs/2003.04297}
\BIBentrySTDinterwordspacing

\bibitem{cl2}
T.~Chen, S.~Kornblith, M.~Norouzi, and G.~Hinton, ``A simple framework for
  contrastive learning of visual representations,'' in \emph{Proc. Int. Conf.
  Mach. Learn.}, 2020, pp. 1597--1607.

\bibitem{cl3byol}
J.-B. Grill, F.~Strub, F.~Altch{\'e}, C.~Tallec, P.~Richemond, E.~Buchatskaya,
  C.~Doersch, B.~Avila~Pires, Z.~Guo, M.~Gheshlaghi~Azar \emph{et~al.},
  ``Bootstrap your own latent-a new approach to self-supervised learning,''
  \emph{Proc. Adv. Neural Inf. Process. Syst.}, vol.~33, pp. 21\,271--21\,284,
  2020.

\bibitem{simclr}
T.~Chen, S.~Kornblith, M.~Norouzi, and G.~Hinton, ``A simple framework for
  contrastive learning of visual representations,'' in \emph{Proc. Int. Conf.
  Mach. Learn.}\hskip 1em plus 0.5em minus 0.4em\relax PMLR, 2020, pp.
  1597--1607.

\bibitem{sim2}
T.~Chen, S.~Kornblith, K.~Swersky, M.~Norouzi, and G.~E. Hinton, ``Big
  self-supervised models are strong semi-supervised learners,'' \emph{Proc.
  Adv. Neural Inf. Process. Syst.}, vol.~33, pp. 22\,243--22\,255, 2020.

\bibitem{Tobler}
W.~R. Tobler, ``\BIBforeignlanguage{en-US}{A computer movie simulating urban
  growth in the detroit region},'' \emph{\BIBforeignlanguage{en-US}{Econ.
  Geogr.}}, p. 234, Jun 1970.

\bibitem{false}
Z.~Zhang, X.~Wang, X.~Mei, C.~Tao, and H.~Li, ``False: False negative samples
  aware contrastive learning for semantic segmentation of high-resolution
  remote sensing image,'' \emph{IEEE Geosci. Remote Sens. Lett.}, vol.~19, pp.
  1--5, 2022.

\bibitem{oscd}
R.~C. Daudt, B.~Le~Saux, A.~Boulch, and Y.~Gousseau, ``Urban change detection
  for multispectral earth observation using convolutional neural networks,'' in
  \emph{Proc. IEEE Int. Geosci. Remote Sens. Symp. (IGARSS)}.\hskip 1em plus
  0.5em minus 0.4em\relax Ieee, 2018, pp. 2115--2118.

\bibitem{levir}
H.~Chen and Z.~Shi, ``A spatial-temporal attention-based method and a new
  dataset for remote sensing image change detection,'' \emph{Remote Sens.},
  vol.~12, no.~10, 2020.

\bibitem{bigearthnet}
G.~Sumbul, M.~Charfuelan, B.~Demir, and V.~Markl, ``Bigearthnet: A large-scale
  benchmark archive for remote sensing image understanding,'' in \emph{Proc.
  IEEE Int. Geosci. Remote Sens. Symp. (IGARSS)}.\hskip 1em plus 0.5em minus
  0.4em\relax IEEE, 2019, pp. 5901--5904.

\bibitem{EuroSAT}
P.~Helber, B.~Bischke, A.~Dengel, and D.~Borth, ``Eurosat: A novel dataset and
  deep learning benchmark for land use and land cover classification,''
  \emph{IEEE J. Sel. Top. Appl. Earth. Obs. Remote Sens.}, 2017.

\bibitem{diffzongshu}
\BIBentryALTinterwordspacing
L.~Yang, Z.~Zhang, Y.~Song, S.~Hong, R.~Xu, Y.~Zhao, Y.~Shao, W.~Zhang, B.~Cui,
  and M.-H. Yang, ``Diffusion models: A comprehensive survey of methods and
  applications,'' \emph{arXiv}, 2022. [Online]. Available:
  \url{https://arxiv.org/abs/2209.00796}
\BIBentrySTDinterwordspacing

\bibitem{diff}
J.~Ho, A.~Jain, and P.~Abbeel, ``Denoising diffusion probabilistic models,''
  \emph{Proc. Adv. Neural Inf. Process. Syst.}, vol.~33, pp. 6840--6851, 2020.

\bibitem{diffgene1}
\BIBentryALTinterwordspacing
A.~Nichol, P.~Dhariwal, A.~Ramesh, P.~Shyam, P.~Mishkin, B.~McGrew,
  I.~Sutskever, and M.~Chen, ``Glide: Towards photorealistic image generation
  and editing with text-guided diffusion models,'' \emph{arXiv}, 2021.
  [Online]. Available: \url{https://arxiv.org/abs/2112.10741}
\BIBentrySTDinterwordspacing

\bibitem{diffgene2}
P.~Dhariwal and A.~Nichol, ``Diffusion models beat gans on image synthesis,''
  \emph{Proc. Adv. Neural Inf. Process. Syst.}, vol.~34, pp. 8780--8794, 2021.

\bibitem{ramesh2022hierarchical}
\BIBentryALTinterwordspacing
A.~Ramesh, P.~Dhariwal, A.~Nichol, C.~Chu, and M.~Chen, ``Hierarchical
  text-conditional image generation with clip latents,'' \emph{arXiv}, vol.~1,
  no.~2, p.~3, 2022. [Online]. Available:
  \url{https://arxiv.org/abs/2204.06125}
\BIBentrySTDinterwordspacing

\bibitem{d}
\BIBentryALTinterwordspacing
W.~Xiang, H.~Yang, D.~Huang, and Y.~Wang, ``Denoising diffusion autoencoders
  are unified self-supervised learners,'' \emph{arXiv}, 2023. [Online].
  Available: \url{https://arxiv.org/abs/2303.09769}
\BIBentrySTDinterwordspacing

\bibitem{ddpm}
\BIBentryALTinterwordspacing
W.~G.~C. Bandara, N.~G. Nair, and V.~M. Patel, ``Ddpm-cd: Remote sensing change
  detection using denoising diffusion probabilistic models,'' \emph{arXiv},
  2022. [Online]. Available: \url{https://arxiv.org/abs/2206.11892}
\BIBentrySTDinterwordspacing

\bibitem{mabo}
\BIBentryALTinterwordspacing
J.~Ma, W.~Xie, Y.~Li, and L.~Fang, ``Bsdm: Background suppression diffusion
  model for hyperspectral anomaly detection,'' \emph{arXiv}, 2023. [Online].
  Available: \url{https://arxiv.org/abs/2307.09861}
\BIBentrySTDinterwordspacing

\bibitem{hadsell}
R.~Hadsell, S.~Chopra, and Y.~LeCun, ``Dimensionality reduction by learning an
  invariant mapping,'' in \emph{Proc. IEEE Conf. Comput. Vis. Pattern Recog.
  (CVPR)}, vol.~2.\hskip 1em plus 0.5em minus 0.4em\relax IEEE, 2006, pp.
  1735--1742.

\bibitem{cmc}
Y.~Tian, D.~Krishnan, and P.~Isola, ``Contrastive multiview coding,'' in
  \emph{Proc. Eur. Conf. Comput. Vis. (ECCV)}.\hskip 1em plus 0.5em minus
  0.4em\relax Springer, 2020, pp. 776--794.

\bibitem{simsiam}
X.~Chen and K.~He, ``Exploring simple siamese representation learning,'' in
  \emph{Proc. IEEE Conf. Comput. Vis. Pattern Recog. (CVPR)}, 2021, pp.
  15\,750--15\,758.

\bibitem{cl4barlow}
J.~Zbontar, L.~Jing, I.~Misra, Y.~LeCun, and S.~Deny, ``Barlow twins:
  Self-supervised learning via redundancy reduction,'' in \emph{Proc. Int.
  Conf. Mach. Learn.}, 2021, pp. 12\,310--12\,320.

\bibitem{ascl}
C.~Feng and I.~Patras, ``Adaptive soft contrastive learning,'' in \emph{Proc.
  Int. Conf. on Pattern Recog.}\hskip 1em plus 0.5em minus 0.4em\relax IEEE,
  2022, pp. 2721--2727.

\bibitem{sen12ms}
\BIBentryALTinterwordspacing
M.~Schmitt, L.~H. Hughes, C.~Qiu, and X.~X. Zhu, ``Sen12ms--a curated dataset
  of georeferenced multi-spectral sentinel-1/2 imagery for deep learning and
  data fusion,'' \emph{arXiv}, 2019. [Online]. Available:
  \url{https://arxiv.org/abs/1906.07789}
\BIBentrySTDinterwordspacing

\bibitem{fmow}
G.~Christie, N.~Fendley, J.~Wilson, and R.~Mukherjee, ``Functional map of the
  world,'' in \emph{Proc. IEEE Conf. Comput. Vis. Pattern Recog. (CVPR)}, 2018,
  pp. 6172--6180.

\bibitem{google}
N.~Gorelick, M.~Hancher, M.~Dixon, S.~Ilyushchenko, D.~Thau, and R.~Moore,
  ``Google earth engine: Planetary-scale geospatial analysis for everyone,''
  \emph{Remote sens. Environ.}, vol. 202, pp. 18--27, 2017.

\bibitem{geossl}
K.~Ayush, B.~Uzkent, C.~Meng, K.~Tanmay, M.~Burke, D.~Lobell, and S.~Ermon,
  ``Geography-aware self-supervised learning,'' in \emph{Proc. IEEE/CVF Int.
  Conf. Comput. Vis. (ICCV)}, 2021, pp. 10\,181--10\,190.

\bibitem{saumoco}
J.~Kang, R.~Fernandez-Beltran, P.~Duan, S.~Liu, and A.~J. Plaza, ``Deep
  unsupervised embedding for remotely sensed images based on spatially
  augmented momentum contrast,'' \emph{IEEE Trans. on Geosci. and Remote
  Sens.}, vol.~59, no.~3, pp. 2598--2610, 2021.

\bibitem{seco}
O.~Manas, A.~Lacoste, X.~Gir{\'o}-i Nieto, D.~Vazquez, and P.~Rodriguez,
  ``Seasonal contrast: Unsupervised pre-training from uncurated remote sensing
  data,'' in \emph{Proc. IEEE/CVF Int. Conf. Comput. Vis. (ICCV)}, 2021, pp.
  9414--9423.

\bibitem{sticl}
H.~Huang, Z.~Mou, Y.~Li, Q.~Li, J.~Chen, and H.~Li, ``Spatial-temporal
  invariant contrastive learning for remote sensing scene classification,''
  \emph{IEEE Geosci. Remote Sens. Lett.}, vol.~19, pp. 1--5, 2022.

\bibitem{s2}
M.~Drusch, U.~Del~Bello, S.~Carlier, O.~Colin, V.~Fernandez, F.~Gascon,
  B.~Hoersch, C.~Isola, P.~Laberinti, P.~Martimort \emph{et~al.}, ``Sentinel-2:
  Esa's optical high-resolution mission for gmes operational services,''
  \emph{Remote sens. Environ.}, vol. 120, pp. 25--36, 2012.

\bibitem{resnet}
K.~He, X.~Zhang, S.~Ren, and J.~Sun, ``Deep residual learning for image
  recognition,'' in \emph{Proc. IEEE Conf. Comput. Vis. Pattern Recog. (CVPR)},
  2016, pp. 770--778.

\bibitem{adam}
\BIBentryALTinterwordspacing
D.~P. Kingma and J.~Ba, ``Adam: A method for stochastic optimization,''
  \emph{arXiv}, 2014. [Online]. Available:
  \url{https://arxiv.org/abs/1412.6980}
\BIBentrySTDinterwordspacing

\bibitem{dira}
F.~Haghighi, M.~R.~H. Taher, M.~B. Gotway, and J.~Liang, ``Dira:
  Discriminative, restorative, and adversarial learning for self-supervised
  medical image analysis,'' in \emph{Proc. IEEE/CVF Int. Conf. Comput. Vis.
  (ICCV)}, 2022, pp. 20\,824--20\,834.

\bibitem{tri}
Q.~Zhang, Y.~Wang, and Y.~Wang, ``Identifiable contrastive learning with
  automatic feature importance discovery,'' in \emph{Proc. Adv. Neural Inf.
  Process. Syst.}, 2023.

\bibitem{unet}
O.~Ronneberger, P.~Fischer, and T.~Brox, ``\BIBforeignlanguage{en-US}{U-net:
  Convolutional networks for biomedical image segmentation},'' in
  \emph{\BIBforeignlanguage{en-US}{Proc. Int. Conf. Med. Image Comput.
  Comput.-Assisted Intervention}}, Jan 2015, p. 234–241.

\bibitem{neumann2019domain}
\BIBentryALTinterwordspacing
M.~Neumann, A.~S. Pinto, X.~Zhai, and N.~Houlsby, ``In-domain representation
  learning for remote sensing,'' \emph{arXiv}, 2019. [Online]. Available:
  \url{https://arxiv.org/abs/1911.06721}
\BIBentrySTDinterwordspacing

\end{thebibliography}
}

\end{document}